\documentclass{article}

\usepackage[numbers]{natbib}

\usepackage[preprint]{neurips_2022}

\usepackage[utf8]{inputenc} % allow utf-8 input
\usepackage[T1]{fontenc}    % use 8-bit T1 fonts
\usepackage{hyperref}       % hyperlinks
\usepackage{url}            % simple URL typesetting
\usepackage{booktabs}       % professional-quality tables
\usepackage{amsfonts}       % blackboard math symbols
\usepackage{nicefrac}       % compact symbols for 1/2, etc.
\usepackage{microtype}      % microtypography
\usepackage{xcolor}         % colors

\usepackage{algorithm}
\usepackage[noend]{algpseudocode}
\usepackage{graphicx}
\usepackage{caption}
\usepackage{subcaption}
\usepackage{hyperref}

\usepackage{array}
\usepackage{amssymb}
\usepackage{amsmath}
\usepackage{xspace}
\usepackage{fancyhdr}
\usepackage{comment}
\usepackage{tabularx}
\usepackage[mathscr]{eucal}
\usepackage{dcolumn}
\usepackage{multirow}
\usepackage{adjustbox}
\usepackage{wrapfig}
\usepackage{array}

\newcolumntype{P}[1]{>{\centering\arraybackslash}p{#1}}
\title{GPPF: A General Perception Pre-training Framework via Sparsely Activated Multi-Task Learning}

\author{%
	Benyuan Sun \quad  Jin Dai  \quad Zihao Liang  \quad Congying Liu \quad Yi Yang \quad Bo Bai \\
	Media Technology Lab, Huawei\\
	\texttt{\{sunbenyuan,daijin7,liangzihao1,liucongying,yangyi16,baibo3\}@huawei.com} \\
}

\begin{document}

\maketitle

\begin{abstract}
  Pre-training over mixtured multi-task, multi-domain, and multi-modal data  remains an open challenge in vision perception pre-training. In this paper, we propose GPPF, a General Perception Pre-training Framework, that pre-trains a task-level dynamic network, which is composed by knowledge "legos" in each layers, on labeled multi-task and multi-domain datasets.
  By inspecting humans'  innate ability  to learn in complex environment, we recognize and transfer three critical  elements to deep networks: (1) simultaneous exposure to diverse cross-task and cross-domain information in each batch. (2) partitioned knowledge storage in separate lego units driven by knowledge sharing. (3) sparse activation of a subset of lego units for both pre-training  and downstream tasks. 
  Noteworthy, the joint training of disparate vision tasks is non-trivial due to their differences in input shapes, loss functions, output formats, data distributions, etc. Therefore, we innovatively develop a plug-and-play multi-task training algorithm, which supports Single Iteration Multiple Tasks (SIMT) concurrently training. SIMT lays the foundation of pre-training with large-scale multi-task multi-domain datasets  and is  proved essential for stable training in our GPPF experiments.
  Excitingly, the exhaustive experiments show that, our GPPF-R50 model achieves significant improvements of 2.5-5.8 over a strong baseline of the 8 pre-training tasks in GPPF-15M and harvests a range of SOTAs over the 22 downstream tasks with similar computation budgets. We also validate the generalization ability of GPPF to SOTA vision transformers with consistent improvements. These solid experimental results fully prove  the effective  knowledge learning, storing, sharing, and transfer provided by our novel GPPF framework. 
  
\end{abstract}

\section{Introduction}

Recent works have shown impressive superiority of large-scale pre-trained models in computer vision (CV) \citep{moco2020,simclr2020,ridnik2021imagenet21k,mae2021}, natural language processing (NLP) \citep{bert18,2020t5,gpt-3-2020,palm2022}, and cross-modal applications \citep{clip21,align21, florence21,flamingo2022}. The pre-training benefits from encoding abundant data distributions into a foundation model and transfers the encoded knowledge  by finetuning or prompting \citep{palm2022}. However, unlike tasks in NLP, where the input and output often lie in similar language spaces, vision tasks are less "self-contained". The input distribution (e.g., domain gap, image shape difference) and output format (e.g., categorical vector, bounding box, and pixel-level prediction map) vary greatly between tasks. Therefore, it is generally hard for a single pre-training task to encode all these visual priors and universally adapt to these disparate downstrem tasks. On the other hand, numerous labeled datasets have been collected in this area with the rising of deep learning. With abundant structured knowledge of different tasks, these datasets are expected to provide essential knowledge for pre-training.

\begin{figure}[t]
	\includegraphics[width=\linewidth]{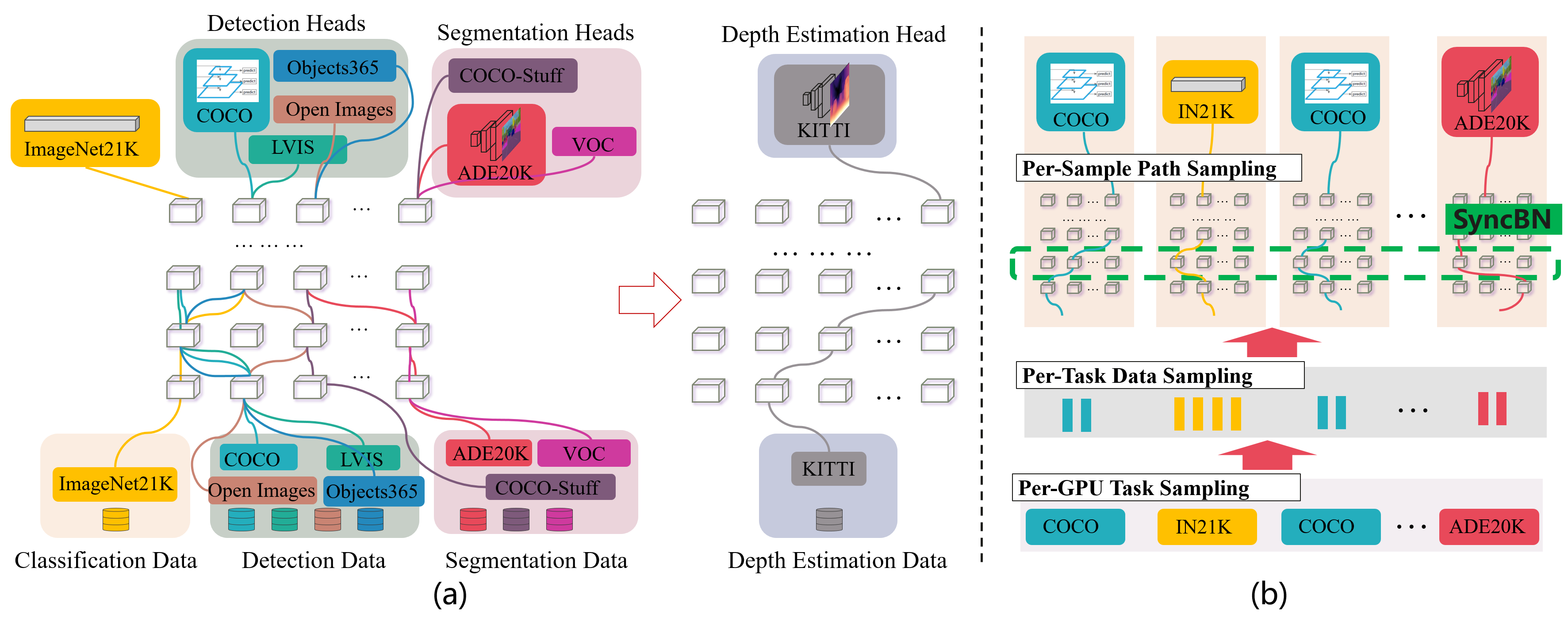}
	\caption{\textbf{The snapshot of GPPF:} (a) During pre-training, GPPF trains a task-level dynamic network on large-scale multi-task and multi-domain datasets. During finefuning, only a sparse routine is selected and finetuned; (b) Single Iteration Multiple Tasks (SIMT) training with cross-task batch normalization. In our experiments, we find SIMT is essential for stable training on large-scale cross dataset training.}
	\label{fig:arch}
	\vspace{-10pt}
\end{figure}

In addition, humans' cognitive flexibility of rapid instructed learning \citep{cole2013rapid} to novel tasks is also found to be related multi-task learning \citep{cole2013multi}. Cole et al. demonstrate that this fast adaptation ability correlates to pattern transfer with compositional coding from diverse learned tasks.  Specifically, when a new task comes, the brain's fronto-parietal control network will sparsely select and reuse a subset of existing connectivity patterns, which keep consistent during new task learning. 
To facilitate a similar knowledge transfer mechanism in deep learning, we propose a  pre-training framework, as depicted in Figure.\ref{fig:arch}, which supports: (1) encoding knowledge from diverse data distributions and annotations simultaneously, by a plug-and-play multi-task multi-domain training algorithm; (2) clustering knowledge into lego units to form "foundation" patterns, by a task-level dynamic network; (3) selecting and reusing knowledge adaptively for downstream tasks, by a differentiable  connection controller. We call this framework as General Perception Pre-training Framework (GPPF).

This proposed framework is similar to the recent idea of Pathways \citep{pathwaysblog2021}, which envisions a multi-task, multi-modal, and sparsely activated network. 
However, as mentioned above, vision tasks vary greatly in input shapes, loss functions, output formats, etc. Implementing a plug-and-play multi-task training framework that does not restrict enrolled tasks is non-trivial. To best of our knowledge, we have not discovered a complete implementation of this idea  on large-scale real-world vision datasets. To address this problem, an important feature we propose in GPPF is the Single Iteration Multiple Tasks (SIMT) concurrently  training. This is achieved by training single task within every  single GPU and sampling different tasks on different GPUs. This strategy is further  combined with synchronized batch normalization (SyncBN) to deliver the problem of statistics deviation between domains. Based on the observations in our experiments, SIMT with SyncBN is essential for the convergence of large-scale multi-task and multi-domain pre-training. It also avoids introducing complex asynchronous dataflow like in the  Pathways training platform \citep{pathways_mlsys} and is compatible  with the popular multi-controller based parallelism  in Pytorch/JAX.

We benchmark our GPPF with the widely accepted ResNet-50 \citep{resnet2016he} architecture and also test it on the recent vision transformer \citep{swin2021liu}. Exhaustive experiments show that, our GPPF-R50 model achieves significant improvements of 2.5-5.8 over a strong baseline of the 8 pre-training tasks in GPPF-15M and harvests a range of SOTAs over the 22 downstream tasks with similar computation budgets. Specifically, for the unseen KITTI depth estimation task, we gets the SOTA within all CNN architectures. Most importantly, all of the reporting performance gains are achieved by a single model. 

Our contributions can be summarized as follows: 

\begin{itemize}
	\setlength{\itemsep}{0pt}
	\setlength{\parsep}{0pt}
	\setlength{\parskip}{0pt}
	\item We propose a promising dynamic perception pre-training framework with a lego unit based structure, a plug-and-play multi-task and multi-domain training mechanism, and a task-level dynamic network.
	
	\item We lay the foundation of the joint training of large-scale multi-task training, which input shapes vary from tasks, together with real-world multi-domain training, by our proposed SIMT with perGPU task sampling and cross task BN.	
	
	\item We propose a sparsely finetuning pipeline which supports precise knowledge transfer for downstream tasks by adaptively  selecting a subset of related lego units.
	
	\item We construct a 15 million multi-task and multi-domain dataset, GPPF-15M. By conducting thorough experiments on GPPF-15M, we verify the SOTA performance of the proposed framework.
\end{itemize}

\section{Related Works}
\vspace{-5pt}

\textbf{Self-Supervised Learning} aims to learn visual representations by self-created supervision on un-labeled images. Many pretext tasks, including jigsaw \citep{jigsaw16}, colorization \citep{colorization16}, and rotation  prediction \citep{rotation2019gidaris} have been proposed to achieve the goal. More recently, contrastive learning \citep{moco2020,simclr2020,simclrv2_2020,byol2020,swav2020,barlowtwins2021} has aroused great attention due to its simplicity and strength. However, these methods often require strong augmentation \citep{simclr2020,swav2020} and are discovered to be less effective on several downstream tasks \citep{ZhouBZZM20,detco2021,densecl2021}. Many improvements have been made to enhance the performance of contrastive learning on target tasks \citep{detco2021,densecl2021} or target scenarios \citep{multisiam2021,Chaitanya2020}.  Another trend for visual pre-training is masked prediction \citep{beit2021,zhou2021ibot,mae2021}, which originated from NLP pre-training \citep{bert18,gpt-3-2020}. The masked prediction fits the vision transformer well and is friendly to smaller batch size training.

\textbf{Multi-Modal Pre-Training} An extensive collection of pre-training based on large-scale image-text data  \citep{AlayracRSARFSDZ20, Jain2021, align21, MiechZATLS19, Pham2021, clip21} has been proposed in these few years. Many of these approaches align the visual and text encoder by contrastive learning. The pre-trained models are proved to excel at zero-shot and few-shot learning especially \citep{clip21, align21}. \citet{flamingo2022} propose Flamingo, which accepts arbitrarily interleaved visual data and text as input. Flamingo gets an even stronger few-shot performance than the contrastive based model. The idea of cross-modal pre-training has also been extended to more modalities \citep{unit2021,vatt2021} to form more general pre-training architecture.

\textbf{Supervised Pre-Training} by classification on ImageNet1K \citep{imagenet} or JFT300M \citep{jft300M} is the most general transfer setting on downstream vision tasks. However, recent studies show that ImageNet1K pre-trained weights might not be beneficial \citep{HeGD19} and even harmful \citep{ZophGLCLC020} to the downstream performance. On the other hand, task-oriented \citep{li2019analysis,objects365} pre-training are discovered to outperform classification pre-training on detection and segmentation. GAIA \citep{gaia2021} pre-trains a detection model by training on multiple detection datasets with unified label space given by word2vec \citep{world2vec}. It also implements a dynamic architecture search for downstream tasks. \citet{mdp2021} pre-trains a  semantic segmentation model by unified training on ADE20K \citep{ZhouZPFB017}, COCO-Stuff \citep{cocostuff18caesar} and Pascal VOC \citep{pascal-voc-2012}. Recently, \citet{MustICCV2021} propose to learn general representation by multi-task self-training. However, it requires pseudo-labeling all tasks on a common large dataset.

\textbf{Multi-Task Learning (MTL)} aims to improve the generalization ability by leveraging information in different tasks. It has been widely studied in the context of deep learning \citep{zhang17survey}. One of the main problems in this area is negative transfer when tasks conflict with each other. The transferability of 26 different vision tasks has been thoroughly studied by \citet{ZamirSSGMS18}. To relieve the pressure of negative transfer, a popular approach is to use task-level dynamic network to enable task-specific parameters \citep{mmoe2018,sun2020adashare,dselectk2021,controllable2022}. We use Gumbel-Softmax \citep{MaddisonMT17,jang2017categorical} controller similar to \citet{sun2020adashare} in this work. On the other hand, \textbf{Multi-Domain Learning (MDL)} focuses on leveraging the information from multiple domains instead of multiple tasks. Directly fusing data with domain gaps also results in negative transfer. Similarly, domain-specific networks, including domain-specific BN \citep{BilenV17}, domain-specific adapter \citep{rebuffi17, rebuffi18} and auto-searched dynamic modules \citep{morgado19,zhao20what,Wang2021WACV} are proposed. Recent study also extends MDL to object detection \citep{uodb19}. 

\vspace{-6pt}
\section{Method}
\vspace{-5pt}
\subsection{Dataset Construction}
\vspace{-5pt}
To support our proposed large-scale multi-task and multi-domain pre-training, we construct a 15 million dataset, which covers the commonly attended classification, detection, and segmentation tasks, called GPPF-15M. Specifically, the cleaned version of ImageNet21K \citep{imagenet, ridnik2021imagenet21k}, which removes infrequent classes and resize the original image to 224 resolution, is adopted for the classification task. For detection and instance segmentation, we choose COCO \citep{LinMBHPRDZ14}, LVIS \citep{GuptaDG19}, Objects365v2 \citep{objects365} and Open Images 2019 Detection Challenge \citep{OpenImages}. Finally, for semantic segmentation, we choose ADE20K \citep{ZhouZPFB017}, Pascal VOC 2012 with augmented labels \citep{pascal-voc-2012, HariharanABMM11}, and COCO-Stuff \citep{cocostuff18caesar}. The resulting GPPF-15M spans a wide range of different classes of multiple hierarchies. It contains 11M classification data, 3.6M detection data and 0.15M semantic segmentation data. We believe this severe task data imbalance reflects the real-world data structure since the per-image annotation cost of different tasks varies greatly. 

\vspace{-3pt}
\subsection{Unified Multi-Task Pre-Training}
\vspace{-3pt}
\textbf{Preliminary} For clarity, we denote the collection of tasks as $T$ and its cardinality as $N_T$. In this work, we restrict ourselves to deep multi-task learning and define task $t \in T$ to be a tuple of dataset and algorithm $(D_t, alg_t)$. Each dataset $D_t = \{(x_i, y_i)\}$ consists of images and annotations. The annotation $y_i$ can be a collection of different labels (e.g., a set of bounding boxes and segmentation masks) or just empty. For each algorithm $alg_t$, a data pre-process function $pre_t$ will augment the image $x$ and fed it into the task network $net_t$. We call the shared parts $M=\cap\{net_t: t\in T\}$ of task networks as the "backbone". There is also a task-specific loss function $loss_t$ which is used to compute gradients and update parameters of $net_t$. In all, an $alg_t=(pre_t, net_t, loss_t)$ is a collection of data pre-process, network, and loss function. For multi-task learning, the overall objective $\sum_{t\in T} \sum_{x \in D_t}w_t * loss_t(x)$ is used to update all tasks simultaneously. $w_t$ is the loss weight of each task, and we keep them equal to 1.0 for all tasks in this work.

\begin{figure}[t]
	\includegraphics[width=\linewidth]{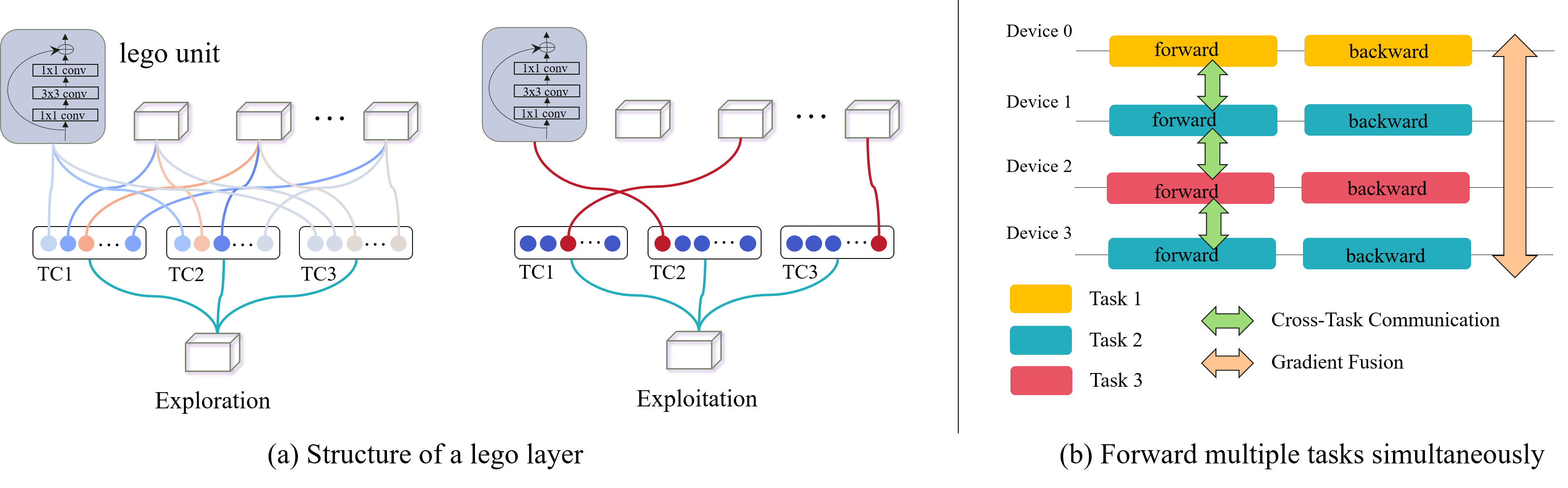}
	\small
	\caption{(a) The lego layer of our task-level dynamic network, which explores possible connections at the start of training and gradually converges to simple unit exploitation. "TC" means task controller. (b) The implemntation of SIMT. Different tasks are sampled on different devices while each device only run single task.}
	\normalsize
	\label{fig:task_gate}
	\vspace{-10pt}
\end{figure}

\textbf{Task-Specific Dynamic Network} As depicted in Figure \ref{fig:arch}, the shared backbone $M$ is constructed by a stack of $L$ lego layers (Figure \ref{fig:task_gate} (a)), which consists of $N$ lego units. In each layer $l$, the $N$ lego units $C_l = \{M_{l, k} | l=1,2,...,L, k=1,2,...,N\}$ have same architecture (e.g., Residual BottleNeck Block \citep{resnet2016he}) but different weights. At the start of the training, each task is densely connected to all the $N$ units. During training, each task explores the best connection pattern guided by its loss function and gradually converges to sparse connection.  We use Gumbel-Softmax \citep{MaddisonMT17,jang2017categorical} to achieve this process of exploration and exploitation. More specifically, for each task $t$, the controller at level $l$ records the probability $p^t_l[k]$ of task $t$ selecting $k^{th}$ candidate unit. During training, the controller explores possible weights by 

\vspace{-20pt}
\small
\begin{align}
u^t_l[k] = {\exp\big((\log p^t_l[k] + G_k) / \tau\big)} / 
{\sum\limits_{i=1}^N\exp\big((\log p^t_l[i] + G_i) / \tau\big)}, k \in {1, ..., N} ~\label{eq:gumbel_softmax}
\end{align}
\normalsize
\vspace{-10pt}

where $G=-\log(-\log U)$ is a standard Gumbel distribution with $U$ sampled i.i.d. from uniform distribution $Unif(0,1)$ and  $\tau$ is the temperature parameter. The temperature decays linearly during the training to ensure the final sampling distribution converges to the original discrete distribution \citep{MaddisonMT17}. During testing, only the unit with largest probality at each layer is activated. Note that  Gumbel-Softmax can sample both hard and soft logits, where the hard logits is obtained with $u^t_{l,hard}[k] - stop\_gradient(u^t_l[k]) + u^t_l[k], u^t_{l,hard}[k] \in \{0, 1\}$, called "hard sample trick". In our experiments, we find that the soft logits is preferable for selection learning, but it costs N times GPU memory than the hard logits. We report the results of soft logits unless specified.

\textbf{Single Iteration Multiple Tasks (SIMT)} The demand for forwarding multiple tasks in a single batch comes from two phases. 
The first is that mismatch between frequently updated backbone parameters and infrequently updated task head parameters leads to unstable training. 
If we only sample one task per batch, some tasks will be updated at least at an interval of $N_T$, which can not be ignored when the number of tasks is large. The other is that MTL algorithm may need to communicate features, statistics, or gradients across tasks. For example, we discover that cross-task communication of BN statistics is important for training backbone with BN. However, this demand can not be trivially implemented in the popular training frameworks since the outputs of different task pre-process $pre_t$ are often inconsistent (e.g., 224 for classification and 1024 for detection). Simultaneously forward inputs with different shapes are not well supported on the current CUDA or architecture.

To address the problem, we keep the program of each GPU to be SIMD, which executes normal single-task training on batched data. But we allow different GPUs to execute different tasks simultaneously  (e.g.,  GPU-0 trains COCO and GPU-1 trains IN21K) (See Figure \ref{fig:task_gate} (b))). More precisely, each GPU is viewed as a virtual container. In each iteration, each virtual device $i$ first samples a task $t_i$ and a batch of corresponding training data $b_i$. Then, the batched data $b_i=(x_i, y_i)$ is used to compute task loss $l_i=loss_{t_i}(net_{t_i}(x_i),  y_i)$. Cross-task feature and statistics communication is done by cross-device communication. In our case, we use SyncBN on all devices to collect the cross-task BN statistics. Finally, the gradient $g_i$ of each GPU is calculated by the back-propagation. To fuse the gradients of each learnable parameter $p$ on different GPUs, we use a task average computed by $\frac{1}{N_{gpu}^p}\sum_{i=1}^{N_{gpu}^p}{g_i[p]}$, where $N_{gpu}^p$ is the number of GPU that uses parameter $p$. It should be noticed that this is different from the sample average since the batch size of $b_i$ on different GPUs may not be the same. Since the communication of different tasks only happens between devices, the choice of architecture, input shapes, or loss functions of single task can be very flexible. As a result, we increase the possible training tasks per batch to $N_{gpu}$, which can be further increased by virtualizing one GPU to multiple devices with techniques like vGPU \footnote{https://www.nvidia.com/en-us/design-visualization/solutions/multi-virtual-gpus/}.

\vspace{-3pt}
\subsection{Knowledge Transfer by Sparse Finetuning}
\vspace{-3pt}
For a novel downstream task, we hope to transfer the most related information from pre-training tasks by reusing the best fitting candidate units. Two methods are proposed to adaptively select the suited routines for the new task:

\vspace{-2pt}
\textbf{Dynamic Finetuning} is similar to pre-training and uses Gumbel-Softmax to explore the routines automatically. The temperature decay in Gumbel-Softmax is also applied to encourage exploration at the beginning and exploitation at the end. During finetuning, the parameters of the backbone and Gumbel-Softmax are updated together. Finally, after training, units with the highest selection probability at each level are extracted and combined as the final downstream network.

\vspace{-2pt}
\textbf{Fixed Path Finetuning} does not learn the task routine and reuses an existing task routine. One way to get the path is to exploit the path in related tasks (e.g., PASCAL VOC detection uses the path of COCO detection). Another way is to use the path learned in dynamic finetuning. The second way is often called "retrain" in the context of multi-task learning. The fixed path finetuning is almost the same as standard finetuning, except that a standard backbone should be firstly extracted from the dynamic backbone. 

\vspace{-4pt}
\section{Evaluation}
\vspace{-6pt}
\subsection{Experimental Settings}
\vspace{-2pt}

In our experiments, we try to build strong baselines for each task. For classification, we apply a linear layer with global average pooling after the final backbone output. Mixup \citep{mixup2018zhang}, Cutmix \citep{cutmix2019yun}, RandAugment \citep{randaug2020cubuk}, Random Erasing \citep{ra2020zhong}, and Label Smoothing \citep{ls2016szegedy} are used. The hyperparameter selection is similar to those in ConvNeXt \citep{convnext2022liu}. For detection and instance segmentation, Mask-RCNN with feature pyramid network (FPN) is utilized. We follow the settings of \citep{copypaste2021ghiasi}, which uses four convolutional layers before the final prediction layers in box and mask head. Copy-paste and large-scale jitter augmentation with size 1024 are also applied. For segmentation, we adopt UperNet \citep{xiao2018unified} architecture and follow the training settings in MMSegmetation \citep{mmseg2020}. We experiment with both ResNet-50 \citep{resnet2016he} and Swin-T \citep{swin2021liu}. Please refer to our Appendix for detailed hyper-parameter and architecture selection. The single task baselines use the same training techniques as the multi-task training except for the training time. The total step for the single task  baseline is selected such that the total training budget of multi-task pre-training is close to the sum of the budget cost by single tasks.

\subsection{Joint Pre-Training}
\vspace{-3pt}

Since for every single task, we only use the standard dataset without introducing extra data like self-training, the additional gain can be attained only with gradient average during multi-task learning. We are eager to find out how the large-scale datasets can benefit from this simple step. The full results of detection, instance segmentation, semantic segmentation, and classification are demonstrated in Table. \ref{table:ablation}. As expected, the joint training gains noticeable performance improvements on all datasets. We demonstrate the detailed increment of each module in the following paragraphs.

\textbf{Importance of SIMT and cross-task BN} In our ResNet-50 experiment, we train the same multi-task setting but with per-batch task sampling strategy, which samples one task per-batch according to the same task sample distribution. The experiment diverges during training and does not have a meaningful result. 
We also experiment with SIMT but without SyncBN. In this case, BN will only calculate local statistics in a single task instead of cross-task mean and variance. Training in this setting also diverges. One reason for this is that different tasks use dissimilar data augmentations and input shapes, which leads to the  difference in BN mean and variance between tasks. 
Throughout our experiments, we observe that SIMT with SyncBN (for backbone with BN) is essential for stable multi-task learning. 

\vspace{-16pt}

\begin{table}[H]
	\centering
	\setlength\tabcolsep{4pt} 
	\setlength{\extrarowheight}{2.3pt}
	\caption{GPPF results on pre-training tasks}
	\label{table:ablation}
	\small
	\resizebox{\textwidth}{!}{
		\begin{tabular}{|l|P{1cm}|P{1cm}|P{1cm}|c|c|c|c|c|c|c|c|c|c|}
			\hline
			\multirow{3}{*}{Method}& \multirow{3}{*}{SIMT} & \multirow{3}{*}{SyncBN} & \multirow{3}{*}{DyNet} &\multicolumn{6}{c|}{Detection \& Instance Segmentation} & \multicolumn{3}{c|}{Semantic Segmentation} & Cls \\
			\cline{5-14}
			&&&& \multicolumn{2}{c|}{COCO} & \multicolumn{2}{c|}{LVIS} & Obj365 & Open & ADE & VOC & COCO-S & IN21K\\
			\cline{5-14}
			&&&&  $AP^{box}$ & $AP^{mask}$  & $AP^{box}$ & $AP^{mask}$ & AP & AP50 & mIOU & mIOU & mIOU & acc-1\\
			\hline
			Single &&&& 46.8 & 42.0 & 28.2 & 26.6 & 24.0 & 52.4 &  42.1  &  76.4  & 40.4 & 35.8 \\
			\hline
			\multirow{5}{*}{GPPF-R50} &&&& \multicolumn{10}{c|}{not converge} \\
			\cline{2-14}
			& \checkmark &&& \multicolumn{10}{c|}{not converge} \\
			\cline{2-14}
			 & \checkmark & \checkmark && 48.8 & 43.8 & 32.2 & 31.2 & 25.5 & 57.6 & 47.2 & 81.9 & 44.5 & 40.2 \\
			\cline{2-14}       
			 & \checkmark &\checkmark&\checkmark(N=2)& 49.7 & 44.5 & 34.0 & 32.1 & 26.6 & 57.7 & 45.3  &  82.2  &  44.0 & 40.6 \\
			\cline{2-14}
			&\checkmark&\checkmark&\checkmark(N=3)& 49.1 & 44.0 & 33.3 & 31.4 & 25.7 & 57.1 & 46.0 & 80.8 & 43.8 & 40.6 \\
			\hline
			GPPF-SwinT &\checkmark&\checkmark&\checkmark(N=2)& 52.1 & 45.3 & 33.3 & 30.5 & 31.6 & 55.9 & 45.8 & 79.7 & 44.8 & 40.7 \\
			\hline
		\end{tabular}
	}
	\normalsize	
\end{table}	

\vspace{-10pt}

\textbf{Performance Gain by Joint Training} Compared to single task training, we obtain significant performance enhancement on all datasets, with cardinality spanning from 20K to 11M. For  object \textbf{detection} tasks, our static GPPF-R50 with SIMT already improves box AP or AP50 on COCO, LVIS, Objects365, and Open Images by 2.0, 2.0, 1.5, and 5.2, respectively. The dynamic backbone further increases the gap to 2.9, 2.5, 2.6, and 5.3, respectively. The improvements in \textbf{instance segmentation} and \textbf{semantic segmentation} are more prominent. Our dynamic network improves COCO and LVIS mask AP by 2.5 and 3.5. It also surpasses single task on ADE20K, Pascal VOC, and COCO-Stuff by 3.2, 5.8, and 3.6, respectively. On the ImageNet21K \textbf{classification}, our joint training with static and dynamic network beats single task by 4.4 and 4.8 on top-1 accuracy.

\begin{figure}[H]
	\vspace{-6pt}
	\includegraphics[width=\linewidth]{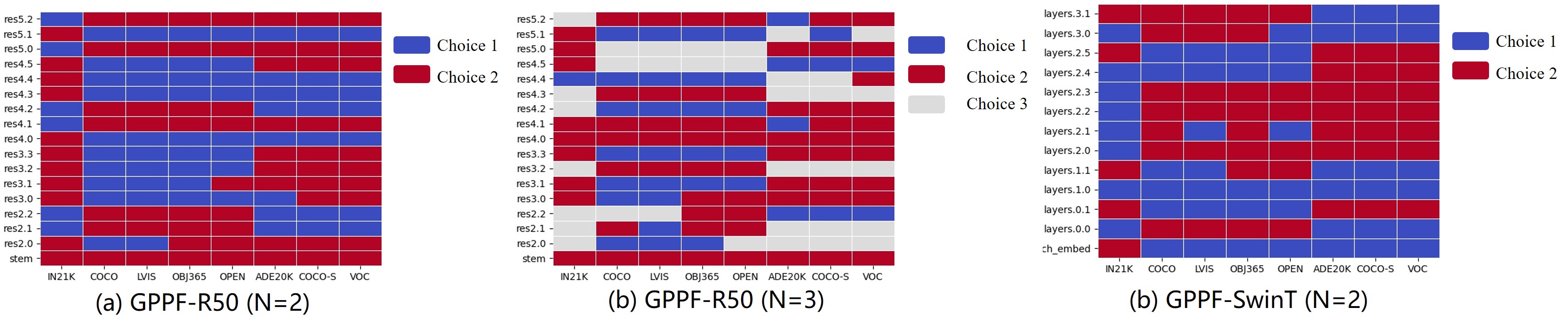}
	\caption{Block Selection of Different Tasks}
	\label{fig:choice}
	\vspace{-15pt}
\end{figure}

\textbf{Performance Gain by Dynamic Network} Compared to static GPPF-R50 network, introducing task-level dynamic network brings better performance. There is an average 0.98, 0.8, and 0.4 improvement on detection, instance segmentation, and classification. Only two tasks of semantic segmentation underperform  the static GPPF-R50. We hypothesize that the phenomenon  is related to dataset size. For the small segmentation datasets, sharing with both classification and segmentation tasks may be more beneficial. However, the classification and detection tasks select different paths in the learning process (See Figure \ref{fig:choice}), making it impossible for semantic segmentation to share with both tasks simultaneously. Increasing $N$ to 3 does not get a further improvement. There are two reasons for this: (1) the network spends more time in exploration and converges slower. (2) Figure \ref{fig:choice} (b) shows that even with 3 candidates  per-layer, most layers do  not activate more than 2 units in the automatic learning. This implies that $N=2$ is nearly sufficient in the current experimental setting. We report the results of $N=2$ as defaults in the following experiments.

\begin{table}[t]
	\centering
	\small
	\resizebox{.95 \textwidth}{!}{
		\begin{minipage}{\textwidth}
			\caption{Comparison on Detection and Instance Segmentation. We compare with models trained both from scratch and pretrained models on COCO and LVIS. On LVIS dataset, we report the two-stage training \citep{copypaste2021ghiasi} results  by further finetuning the final classification layer with Class Balanced Loss \citep{cui2019balance}.}
			\label{table:detection}
			\resizebox{\textwidth}{!}{
				\begin{tabular}{l|c|c|c|cc|cc}
					\toprule
					Method& Backbone & Architecture & \#Param & $AP^{box}_{coco}$ & $AP^{mask}_{coco}$ & $AP^{box}_{lvis}$ & $AP^{mask}_{lvis}$ \\
					\hline
					\multicolumn{8}{c}{\scriptsize{Scratch}} \\
					%% Single & R50-MR & 46M & 46.8 & 42.0 & 28.2 & 26.6 & 24.0 & 52.4 \\
					Copy-Paste \citep{copypaste2021ghiasi} & ResNet-50 & Mask-RCNN & 24M & 48.2 & 42.4 & 34.2 & 32.3\\
					Copy-Paste \citep{copypaste2021ghiasi} & ResNet-101 & Mask-RCNN & 46M & 49.8 & 43.6 & 36.4 & 34.0 \\
					\hline
					\multicolumn{8}{c}{\scriptsize{ImageNet1K Classification Pre-Training}} \\
					Swin-T \citep{swin2021liu} & Swin-T & Mask-RCNN & 29M & 46.0 & 41.6 & / & / \\
					Swin-T \citep{swin2021liu} & Swin-T & Cascade-RCNN & 29M & 50.5 & 43.7 & / & / \\
					% Swin-S \citep{swin2021liu} & Swin-T & Mask-RCNN & 69M & 49.8 & 43.6 & / & / \\
					% Swin-S \citep{swin2021liu} & Swin-T & Cascade-RCNN & 107M & 49.8 & 43.6 & / & / \\
					\hline
					\multicolumn{8}{c}{\scriptsize{ImageNet1K Self-supervised Pre-Training}} \\
					MoCo-v3 \citep{chen2021mocov3}           & ViT-B  & Mask-RCNN & 86M  & 47.9  & 42.7 & / & / \\
					BEiT \citep{beit2021}           & ViT-B  & Mask-RCNN & 86M  & 49.8  & 44.4 & / & / \\
					MAE  \citep{mae2021}       & ViT-B  & Mask-RCNN & 86M  & 50.3  & 44.9 & / & / \\
					\hline
					\multicolumn{8}{c}{\scriptsize{Single/Multiple Datasets Detection Pre-Training}} \\
					Objects365(pretrain) \citep{objects365}      & ResNet-50 & Faster-RCNN  & 24M & 42.3 & / & / & / \\
					GAIA \citep{gaia2021}  & ResNet-50 & Faster-RCNN  & 24M & 44.8 & / & / & / \\
					GAIA-TSAS \citep{gaia2021} & ResNet-50 & Cascade-RCNN & 24M & 49.1 & / & / & / \\
					\midrule
					\multicolumn{8}{c}{\scriptsize{GPPF-15M Multi-Task Pre-Training}} \\  
					GPPF-R50 & ResNet-50 & Mask-RCNN & 24M & 49.7 & 44.5 & 37.4 & 35.9 \\
					GPPF-SwinT & Swin-T & Cascade-RCNN & 29M & 52.1 & 45.3 & 39.6 & 36.1 \\
					\bottomrule
			\end{tabular}}
		\end{minipage}
	}	
	\normalsize
	\vspace{-10pt}
\end{table}

\textbf{Adaptation Ability to SOTA Transformer} To test the generalization ability of GPPF to other network architecture, we experiment with the recent vision transformer Swin-Transformer \citep{swin2021liu}. Specifically, we choose Swin-T, which has a similar model size to ResNet-50. Swin-T uses Layer Normalization (LN) \citep{ln16ba} instead of BN. Therefore, we do not introduce any cross GPU synchronization in the normalization layers. The experimental results demonstrate that GPPF suits vision transformers well. By joint training on the same datasets, GPPF-SwinT gets box-AP and mask-AP of 52.1 and 45.3 on COCO dataset (See Table \ref{table:detection}), which surpasses the results reported in the original paper by 1.6 and 1.6 with a clear margin. On other datasets, GPPF-SwinT also gets strongly competitive performance. This proves the generalization  ability of GPPF to some extent. Another direction is to prove the generaliztion ability of GPPF to large models. Due to computation resource limitations, this is left as future work.

\textbf{Task Selection in Dynamic Network.} We visualize the task grouping of pre-train networks in Figure \ref{fig:choice}. First, without any restrictions, similar tasks still group together. In GPPF-R50 with $N=2$, the four detection tasks share almost all blocks. In the $N=3$ case, the detection tasks only differs in bottom blocks but share all the top blocks. This obeys the intuition that detection tasks share similar high-level knowledge but may differ in low-level data distribution. Second, different groups of tasks only share a smaller subset of lego units. The classification and detection in GPPF-R50(N=2) almost selects different paths, while segmentation shares information with both groups of tasks. Third, in the $N=3$ case, different tasks within the same group does not diverge to different branches, even though "blank" units are available. This suggest sharing knowledge is a better choice than training all units alone.

\begin{wraptable}{r}{0.65\textwidth}
	% \begin{table}[t]
		\centering
		\small
		\vspace{-0.35cm}
		\caption{Semantic Segmentation: measured in mIOU. We report results without test-time augmentation (TTA).}
		\vspace{-0.1cm}
		\resizebox{0.62\textwidth}{!}{
			\begin{tabular}{l|c|c|c|cc}
				\toprule
				\label{table:segmentation}
				Method & Backbone & Architecture & \#Param & ADE20K & VOC \\
				\hline
				\multicolumn{6}{c}{\scriptsize{ImageNet1K Classification Pre-Training}} \\
				ResNet50                     & ResNet-50 & UperNet & 24M & 42.1 & 76.4\\
				Swin-T \citep{swin2021liu}   & Swin-T &  UperNet & 29M  &  44.5 & / \\
				% HRNetV2 \citep{}             & ResNet-50 & HRNetV2 & 24M & / & 80.3\\
				\hline
				\multicolumn{6}{c}{\scriptsize{ImageNet1K Self-Supervised Pre-Training}} \\
				BYOL \citep{byol2020} & ResNet-50  & FCN & 24M & / & 75.7 \\
				MoCo-v3 \citep{chen2021mocov3}  & ViT-B &  UperNet &  86M   &  47.3  &  /    \\
				BEiT \citep{beit2021}  & ViT-B &  UperNet &  86M   &  47.1  &  /   \\
				MAE \citep{mae2021}  & ViT-B &  UperNet &   86M  &  48.1  &  /    \\
				\hline
				\multicolumn{6}{c}{\scriptsize{Single/Multiple Datasets Segmentation Pre-Training}} \\
				MDP \citep{mdp2021}    & ResNet-50  & DeepLab-v3 &   24M  &  42.7  &  77.8   \\
				\hline
				\multicolumn{6}{c}{\scriptsize{GPPF-15M Multi-Task Pre-Training}} \\
				GPPF-R50   & ResNet-50 &  UperNet &  24M   &  45.3  &  82.2    \\
				GPPF-SwinT   & Swin-T &  UperNet &   29M  &   45.8 &  79.9  \\
				\bottomrule
			\end{tabular}
			
		}
		\normalsize
		% \end{table}
	\vspace{-0.23cm}
\end{wraptable}

\textbf{Comparison with SOTA} We compare our methods to SOTA with similar computation budget and model size. \textbf{Detection and instance segmentation} are presented in Table \ref{table:detection}. Our GPPF with ResNet-50 (GPPF-R50) surpasses all methods using same architecture with large margin. Specifically, Copy-Paste \citep{copypaste2021ghiasi} is an extreme strong baseline with 400 epoch training, where ImageNet pre-train is discovered harmful in this long-time training setting. Our GPPF-R50 still outperforms it on COCO box AP, COCO mask AP, LVIS box AP, LVIS mask AP with 1.6, 2.1, 3.2, 3.6, respectively. 
GAIA \citep{gaia2021} is a multi-dataset pre-training model obtained by jointly training on several detection datasets. GAIA-TSAS further introduces deformable convolution \citep{dcn2017dai}, cascade head \citep{cascadercnn}, and neural architecture  search \citep{darts19}. Our model still outperforms it without these techniques. Even compared with SOTA architectures, including ViT and Swin-Transformer, the result of GPPF-R50 is still comparable. Interestingly, our joint training benefits more on the long-tailed LVIS dataset. This can be explained by the abundant class and segmentation information provided by other tasks.

The results of \textbf{Semantic Segmentation}  is  presented in Table \ref{table:segmentation}. Similarly, our GPPF-R50 outperforms all ResNet-50 results and is comparable to ViT and Swin-Transformer. MDP\citep{mdp2021} is a segmentation pre-train on several segmentation datasest. However, the total number of segmentation pre-training data is very limited, even with the data fusion. On the other hand, our joint training makes it possible for semantic segmentation to observe knowledge from large-scale classification and detection datasets. As a result, our GPPF presents a clear advantage over MDP.

\vspace{-5pt}
\subsection{Transfer Ability}

\vspace{-5pt}
\subsubsection{Depth Estimation}
\vspace{-5pt}
Depth estimation is an important task in computer vision. However,  the data of depth estimation is often scarce or sparse since an extra capture device like lidar is required. We wish the knowledge like category and contour of common objects can be transferred to depth estimation by our pre-train model. Quantitative results are listed in Table \ref{table:depth}. We choose BTShead \citep{lee2019big} as the task head. Compared with the current SOTAs, our GPPF-R50 surpasses all CNN backbones, including ResNet-101, and ResNeXt-101 and is only slightly  inferior to the Swin-L SOTA. Compared  to other pre-train models, GPPF-R50 shows a clear advantage over them. Qualitatively, the inference results in the Appendix demonstrate our GPPF-R50 has more robust region continuity in semantic areas over the ImageNet pre-training. This also proves the success of transferring semantic information into the downstream tasks. 

\vspace{-5pt}

\begin{table}[H]
	\centering
	%\small
	\vspace{-10pt}
	\caption{Transfer Learning Results on KITTI Depth Estimation}
	\label{table:depth}
	\resizebox{\textwidth}{!}{
		\begin{tabular}{l|c|c|ccccccc}
			\toprule
			\multicolumn{10}{c}{\footnotesize{ImageNet1K/ImageNet21K Classification Pre-Training}} \\
			Method  & backbone & \#Param & d1 & d2 & d3 & AbsRel & SqRel & RMSE & RMSElog\\
			\hline
			baseline \citep{lee2019big} & ResNet-50  & 24M  & 0.954 & 0.992 & 0.998 & 0.061 & 0.250 & 2.803 & 0.098 \\
			% baseline \citep{lee2019big} & R101  & 46.4M  & 0.954 & 0.992 & 0.998 & 0.061 & 0.261 & 2.834 & 0.099 \\
			
			% \citep{}    & DenseNet161  & 26.3M & 0.956 & 0.994 & 0.999 & 0.062 & 0.240 & 2.708 & 0.096 \\
			LapDepth \citep{song2021monocular}            & ResNeXt-101 & 58M & 0.962 & 0.994 & 0.999 & 0.059 & 0.212 & 2.446 &  0.091 \\
			NeWCRFs \citep{yuan2022newcrfs}       & Swin-L & 197M & 0.974 & 0.997 & 0.999 & 0.052 & 0.155 &  2.129 & 0.079 \\
			BinsFormer \citep{binsformer2022}    & Swin-L & 197M & 0.974 & 0.997 & 0.999 & 0.052 & 0.151 & 2.098 &  0.145 \\
			\midrule
			\multicolumn{10}{c}{\footnotesize{Same Architecure \citep{lee2019big} with Different Pre-Training}} \\
			Scratch              & ResNet-50 & 24M & 0.914 & 0.980 & 0.995 & 0.11 & 0.372 & 3.151 &  0.145 \\
			ImageNet         & ResNet-50 & 24M & 0.963 & 0.993 & 0.999 & 0.062 & 0.239 & 2.797 &  0.096 \\
			COCO            & ResNet-50 & 24M & 0.952 & 0.994 & 0.999 & 0.064 & 0.225 & 2.577 &  0.098 \\
			MoCo-v2 \citep{mocov2_2020chen}        & ResNet-50 & 24M & 0.955 & 0.994 & 0.999 & 0.061 & 0.216 & 2.534 & 0.095  \\
			CLIP \citep{clip21}       & ResNet-50 & 24M & 0.925 & 0.986 & 0.997 & 0.078 & 0.327 & 3.091 & 0.121  \\
			% self-supervised \citep{}   & R50  & 23.5M  & 0.947 & 0.993 & 0.998 & 0.071 & 0.261 & 2.674 & 0.104 \\
			\midrule
			\multicolumn{10}{c}{\footnotesize{GPPF-15M Multi-Task Pre-Training}} \\
			GPPF-R50(IN21K)  & ResNet-50 & 24M & 0.957 & 0.994 & 0.999 & 0.060 & 0.202 & 2.438 & 0.094  \\
			GPPF-R50(ADE20K) & ResNet-50 & 24M & 0.961 & 0.995 & 0.999 & 0.058 & 0.192 & 2.388 & 0.091  \\
			GPPF-R50(COCO)              & ResNet-50 & 24M & 0.965 & 0.996 & 0.999 & 0.057 & 0.180 & 2.252 &  0.088 \\
			GPPF-R50(Dyfinetune) & ResNet-50 & 24M & 0.966 & 0.995 & 0.999 & 0.065 & 0.224 & 2.382 & 0.094 \\
			\bottomrule
		\end{tabular}
	}
	\vspace{-13pt}
\end{table}

\vspace{-5pt}
\subsubsection{Detection}
\vspace{-5pt}
Except for COCO and LVIS, which has over 100k images, we also wish to uncover how detection datasets with the smaller dataset can benefit from GPPF. We choose UODB \citep{uodb19} for verification, a collection of 11 detection datasets. Since we have already trained with COCO, it is removed from our experiments. The results are presented in Table \ref{table:detection_uodb}. Compared to the ImageNet pre-trained baseline, we present significant improvements of over 10.9 AP50 on average. On extreme small datasets like Clipart, the improvement increases to over 42 on AP50. Our method also surpasses GAIA \citep{gaia2021}, a multi-dataset pre-trained detection model, and its stronger version, GAIA-TSAS , which further uses stronger techniques as disscussed above. This again proves the advantage of multi-task instead of single task pre-traininig.

\begin{table}[h]
	\centering
	\small
	\setlength\tabcolsep{3pt} 
	\resizebox{\textwidth}{!}{
	\begin{minipage}{\textwidth}
	\caption{Transfer Learning Results on UODB detection: we report AP50 used in Pascal VOC  2012 as previous papers. We follow the settings in GAIA and initialize the detection head parameters with those in the Objects365. But we do not load the final prediction layer by finding the most similar classes in pre-trained model. Baselines all use ImageNet pre-training.}
	\label{table:detection_uodb}
	\resizebox{\textwidth}{!}{
	\begin{tabular}{l|cccccccccc|c}
		\toprule
		Method@AP50     & KITTI & WiderFace & VOC & LISA & DOTA & Watcolor & Clipart & Comic & Kitchen & DLesion & mean \\
		\hline
		number of train sets & 7k & 13k & 16k & 8k & 14k & 1k & 0.5k & 1k  & 5k & 23k & / \\
		domain & traffic & face & natural & traffic & aerial & watercolor & clipart & comic  & indoor & medical & / \\
		\hline
		Baseline \citep{uodb19}             & 64.3 & 48.8 & 78.5 & 88.3 & 57.5 & 52.6 & 31.2 & 45.8 & 87.7 & 51.2 & 60.6 \\
		Baseline(FPN) \citep{gaia2021}      & 67.1 & 62.1 & 81.5 & 90.0 & 68.3 & 53.4 & 31.2 & 45.5 & 89.5 & 57.4 & 64.6 \\
		DA \citep{uodb19}           & 68.0 & 51.3 & 82.4 & 87.6 & 56.3 & 60.6 & 55.8 & 53.4 & 90.0 & 53.4 & 65.9 \\
		GAIA \citep{gaia2021}         & 72.9 & 62.6 & 85.9 & 91.2 & 69.2 & 63.5 & 67.9 & 57.0 & 89.8 & 59.4 & 71.9 \\
		GAIA-TSAS \citep{gaia2021}    & 75.6 & 62.7 & 87.4 & 92.1 & 70.8 & 69.7 & 72.2 & 61.1 & 90.1 & 62.1 & 74.4 \\
		\midrule
		GPPF-R50(IN21K)                 & 71.3 & 65.9 & 87.3 & 96.2 & 72.8 & 56.0 & 68.3 & 55.0 & 95.1 & 62.5 & 73.0 \\
		GPPF-R50(Objects365)                  & 74.2 & 66.5 & 88.4 & 95.9 & 71.7 & 66.4 & 73.6 & 61.1 & 95.8 & 61.4 & 75.5 \\
		GPPF-R50(Dyfinetune)                  & 76.1 & 66.2 & 88.1 & 96.8 & 72.0 & 68.2 & 77.5 & 62.0 & 95.7 & 62.6 & 76.5 \\
		\bottomrule
	\end{tabular}
	}
	\end{minipage}
	}
	\normalsize
	\vspace{-10pt}
\end{table}

\vspace{-3pt}
\subsubsection{Classification}

\vspace{-3pt}
We first compare the finetuning results on ImageNet1K (Table \ref{table:imagenet}). We follow the training techniques stated by \citet{convnext2022liu}, except we do not use AdamW \citep{adamw1029losh}, Stochastic Depth \citep{densenet2017huang}, and EMA \citep{Polyak1992}. GPPF-R50 gains about 1\% improvements in top-1 accuracy compared to our baseline. It also surpasses the R50 trained with more techniques and SupCon \citep{Khosla2020scl}, which combines supervised learning and contrastive  learning. 

\begin{wraptable}{r}{0.4\textwidth}
	\vspace{-22pt}
	\centering
	\setlength\tabcolsep{3.8pt}
	\small
	\caption{ImageNet1K Results}
	\label{table:imagenet}
	\vspace{-5pt}
	% \resizebox{0.25\textwidth}{!}{
		\begin{tabular}{lcc}
			\toprule
			Method & Acc-1 & Acc-5 \\
			\midrule
			Scratch(repro) & 78.12 & 93.95 \\
			R50(all) \citep{convnext2022liu} & 78.8 & / \\
			SupCon \citep{Khosla2020scl} & 78.7  & 94.3 \\
			GPPF-R50(IN21K) & 79.18 & 94.59 \\
			GPPF-R50(Dyfinetune) & 79.04 & 94.44 \\
			\bottomrule
		\end{tabular}
		% }
	\normalsize
	\vspace{-10pt}
\end{wraptable}

Similarly, we also perform finetuning on a series of downstream classification datasets, presented in Table \ref{table:downstream_classification}. GPPF-R50 outperforms previous methods on 7 of the 10 datasets and is only slightly worse on Food101 and Cars. Although  the comparison is not totally fair on the pre-trained data, it is still meaningful to know that these downstream tasks can be further improved by scaling the pre-trained data. Our framework gives a solution to scale with classification data of different shapes and classes, which is important if we want to utilize all existing data.

\begin{table}[H]
	\vspace{-10pt}
	\centering
	\setlength\tabcolsep{3.8pt}
	\small
	\caption{Transfer Learning Results on Classification: we report Top-1 accuracy on Food101, CIFAR10, CIFAR100, SUN397, Cars, DTD and Mean per-class accuracy on Aircraft, Pets, Caltech-101 and Flowers follow the previous papers.}
	\label{table:downstream_classification}
	\resizebox{\textwidth}{!}{
		\begin{tabular}{l c c c c c c c  c c c}
			\toprule
			{ Method} & { Food101} & { CIFAR10} & { CIFAR100}  & { SUN397}  & { Cars} & { Aircraft} & { DTD} & { Pets} & { Caltech-101} & { Flowers} \\
			\midrule
			Random init \citep{simclr2020} & $86.9$ & $95.9$ & $80.2$   & $53.6$ & $91.4$ & $85.9$ &  $64.8$ & $81.5$ & $72.6$ & $92.0$  \\
			ImageNet1K\citep{simclr2020} & $88.3$ & $97.5$ & $86.4$ &  $64.3$ & $92.1$ & $86.0$ & $74.6$ & $92.1$ & $93.3$ & $97.6$ \\
			SimCLR \citep{simclr2020}  & $88.2$ & $97.7$ & $85.9$ &  $63.5$ & $91.3$ & $88.1$ & $73.2$ & $89.2$ & $92.1$ & $97.0$  \\
			BYOL \citep{byol2020} & $88.5$ & $97.8$ & $86.1$ &  $63.7$& $91.6$ & $88.1$ & $76.2$& $91.7$& $93.8$ & $97.0$\\
			GPPF-R50(IN21K) & $87.7$ & $97.9$ & $87.1$ & $64.3$ & $91.8$ & $89.0$ & $77.2$ & $92.3$ & $95.1$ & $98.4$ \\
			\bottomrule
		\end{tabular}
	}
    \normalsize
   \vspace{-8pt}
\end{table}
% \subsection{GPPF on 3D Detection}

\section{Discussion}

\vspace{-10pt}
Note that our framework is fully compatible with self-supervised learning. Recent researches prove that a combination of supervised and self-supervised training \citep{Khosla2020scl} can improve performance. However, self-supervised algorithms are unexcavated in this work, mainly because they are much more resource-consuming and might require strict conditions like large batch training. In expectation, introducing self-supervised tasks with numerous unlabeled data to our GPPF will be beneficial. 
% Another important topic we dicuss in this paper is cross-task BN. Recent SOTA CNN \citep{convnext2022liu} and transformer \citep{swin2021liu} architectures prefers LN than BN, which does not have the problem of distribution shifting. The comparison of BN and LN in multi-domain and multi-task learning is an important future work. 
Another feature we do not include in this work is supporting non-identical network units in each lego layer. Different tasks may benefit from different  architectures  (e.g., deformable convolution \citep{dcn2017dai}) or different depths. The support for these functions is left as future works.

\vspace{-5pt}
\section{Conclusion}
\vspace{-6pt}
In this paper, we propose GPPF, a General Perception Pre-training Framework that  supports versatile multi-task and multi-domain pre-training. To support joint training of vision tasks with disparate input shapes, data distributions, and annotation formats, we propose a flexible plug-and-play multi-task training algorithm with SIMT. We also introduce a task-level dynamic network, which encourages knowledge  sharing and relieves negative transfer. We conduct thorough  experiments on the constructed GPPF-15M dataset and validate our GPPF clearly.
We believe  a flexible plug-and-play multi-task pre-training  framework is essential for general vision pre-training and hope our framework can provide some inspiration for future works. 

\bibliographystyle{abbrvnat}
\bibliography{ref}

\appendix

\section{CKA Analysis on GPPF Joint Training}

\subsection{CKA Analysis}

Centered Kernel Alignment (CKA) \citep{cka2019} is an analysis method for feature similarity of deep networks. It has been widely used to inspect several properties of deep and wide networks \citep{cka2019,widedeep2021,Raghu2021}. Given two feature matrices $X \in \mathbb{R}^{n\times p_1}$ and $Y \in \mathbb{R}^{n\times p_2}$, the similarity is  calculated by:

\begin{equation}
	CKA(X, Y) = \frac{HSIC(K, L)}{\sqrt{HSIC(K, K)HSIC(L, L)}}
\end{equation}

where $K_{ij}=k(x_i, x_j)$, $L_{ij}=l(y_i, y_j)$ and HSIC is Hilbert-Schmidt Independence Criterion. The $k$ and $l$ are two kernels. HSIC can be estimated by $tr(KHLH)/(n-1)^2$, where H is a centering matrix calculated by $H_n = I_n - \mathbf{11}^T/n$. In this paper, we adopt the batched computation of CKA proposed by \citet{widedeep2021}.

\subsection{Single Task Training v.s. GPPF Joint Training}

\begin{figure}[H]
	\centering
	\begin{subfigure}{0.32\linewidth}
		\centering
		\includegraphics[width=\linewidth]{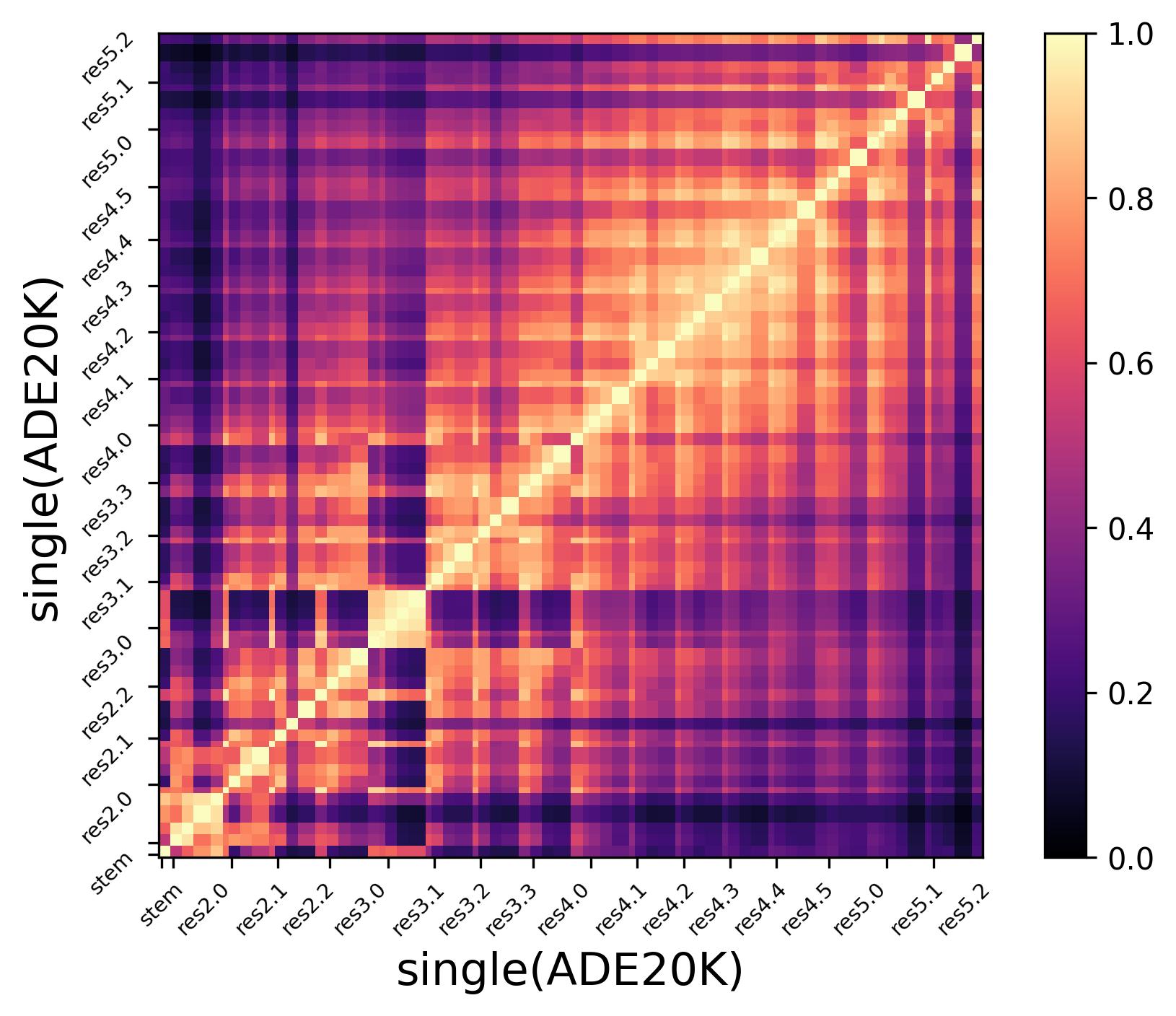}
		\caption{Single}
	\end{subfigure}
	\hfill
	\begin{subfigure}{0.32\linewidth}
		\centering
		\includegraphics[width=\linewidth]{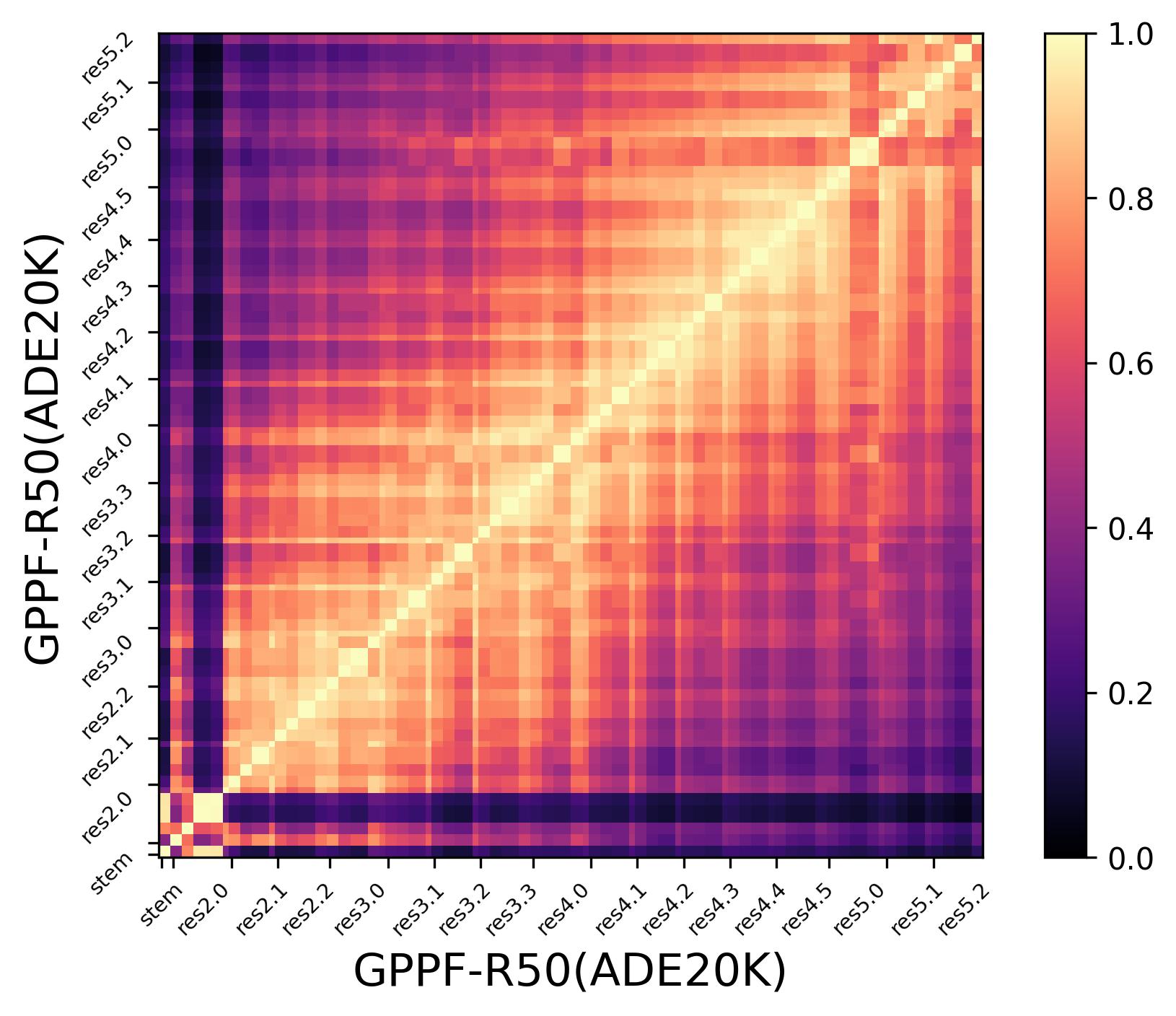}
		\caption{GPPF-R50}
	\end{subfigure}
	\hfill
	\begin{subfigure}{0.32\linewidth}
		\centering
		\includegraphics[width=\linewidth]{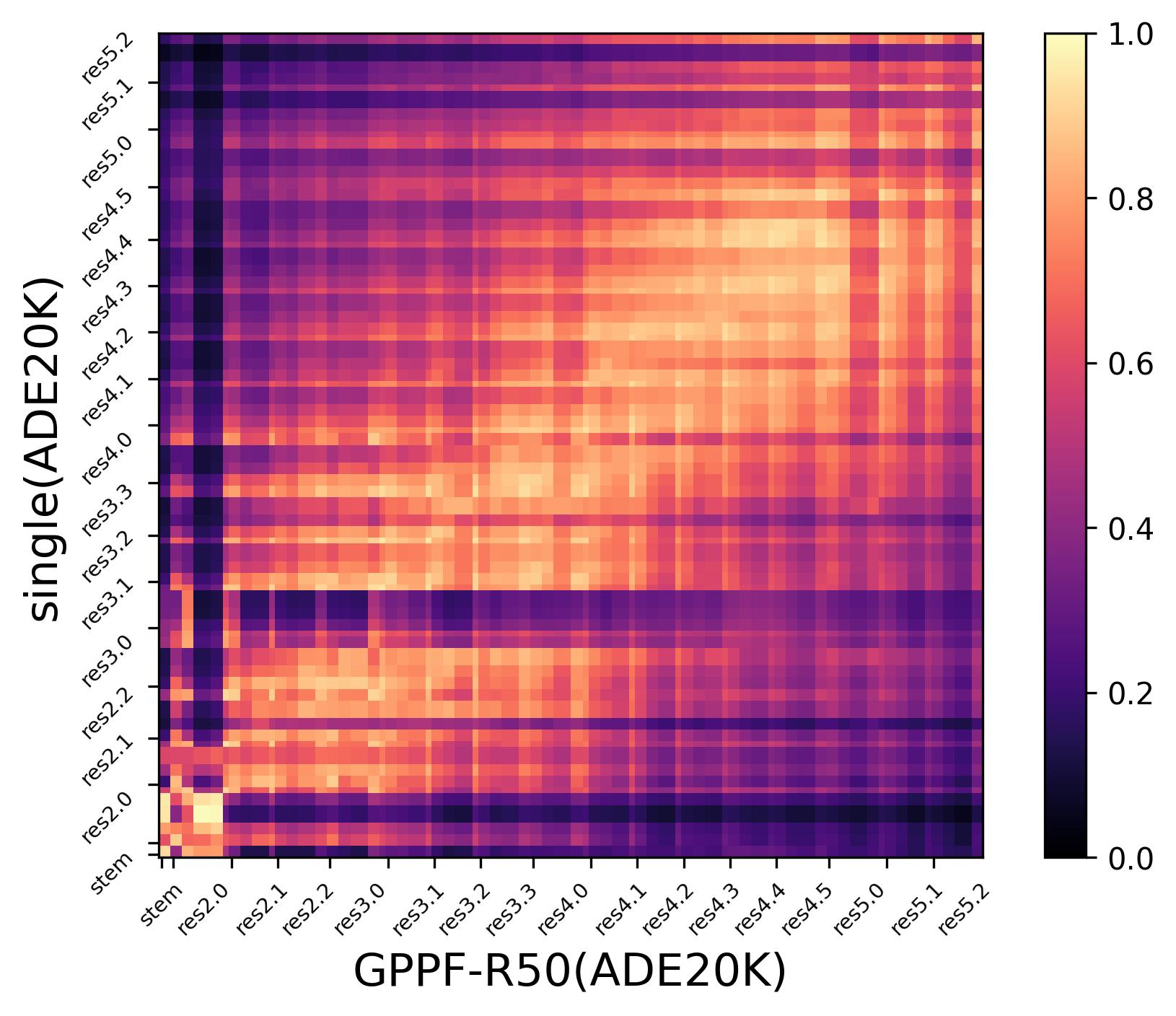}
		\caption{Cross Model Similarity}
	\end{subfigure}
	\caption{CKA analysis on ADE20K segmentation models.}
	\label{fig:ade20k}
\end{figure}

\begin{figure}[H]
	\centering
	\vspace{-15pt}
	\begin{subfigure}{0.32\linewidth}
		\centering
		\includegraphics[width=\linewidth]{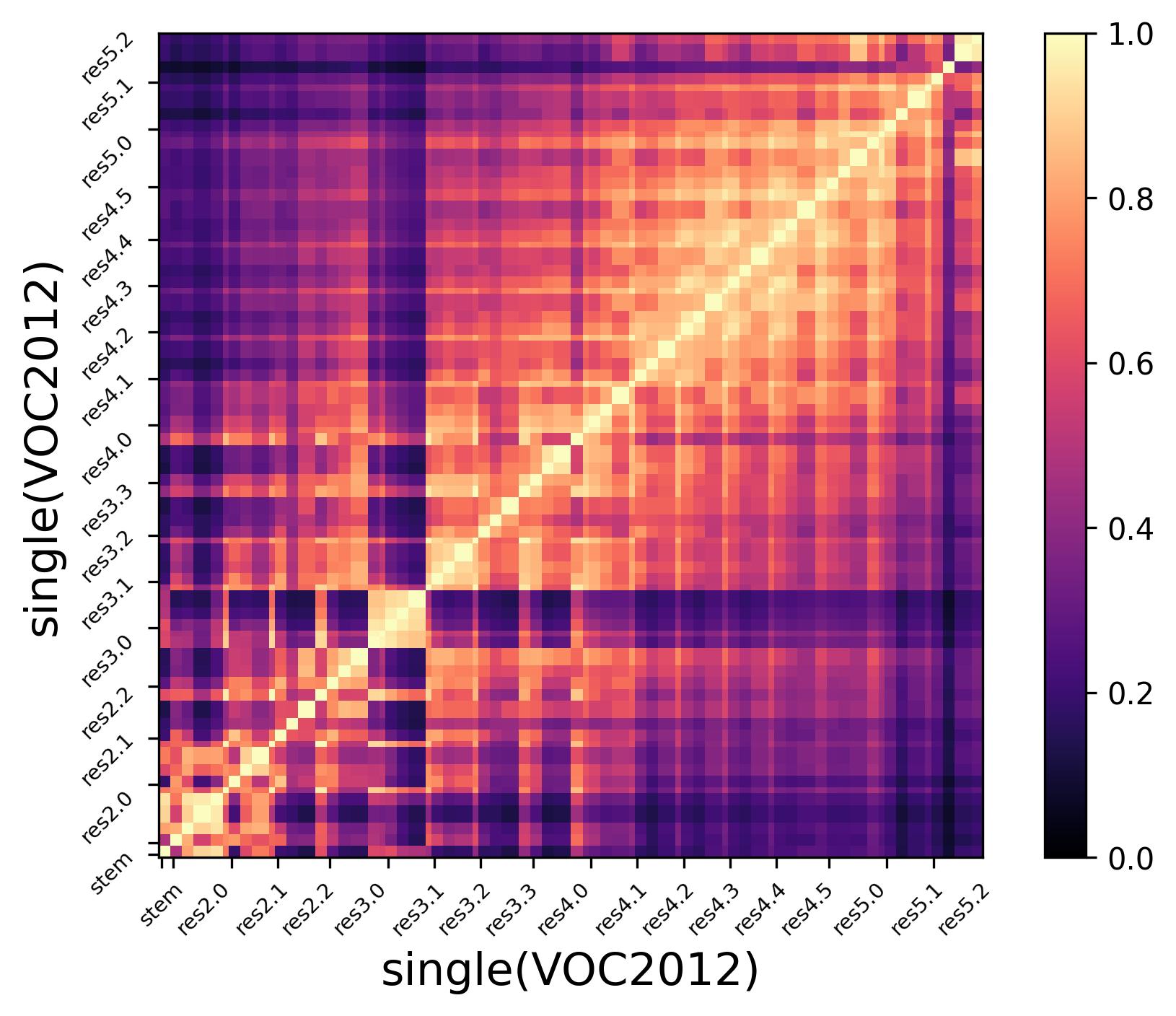}
		\caption{Single}
	\end{subfigure}
	\hfill
	\begin{subfigure}{0.32\linewidth}
		\centering
		\includegraphics[width=\linewidth]{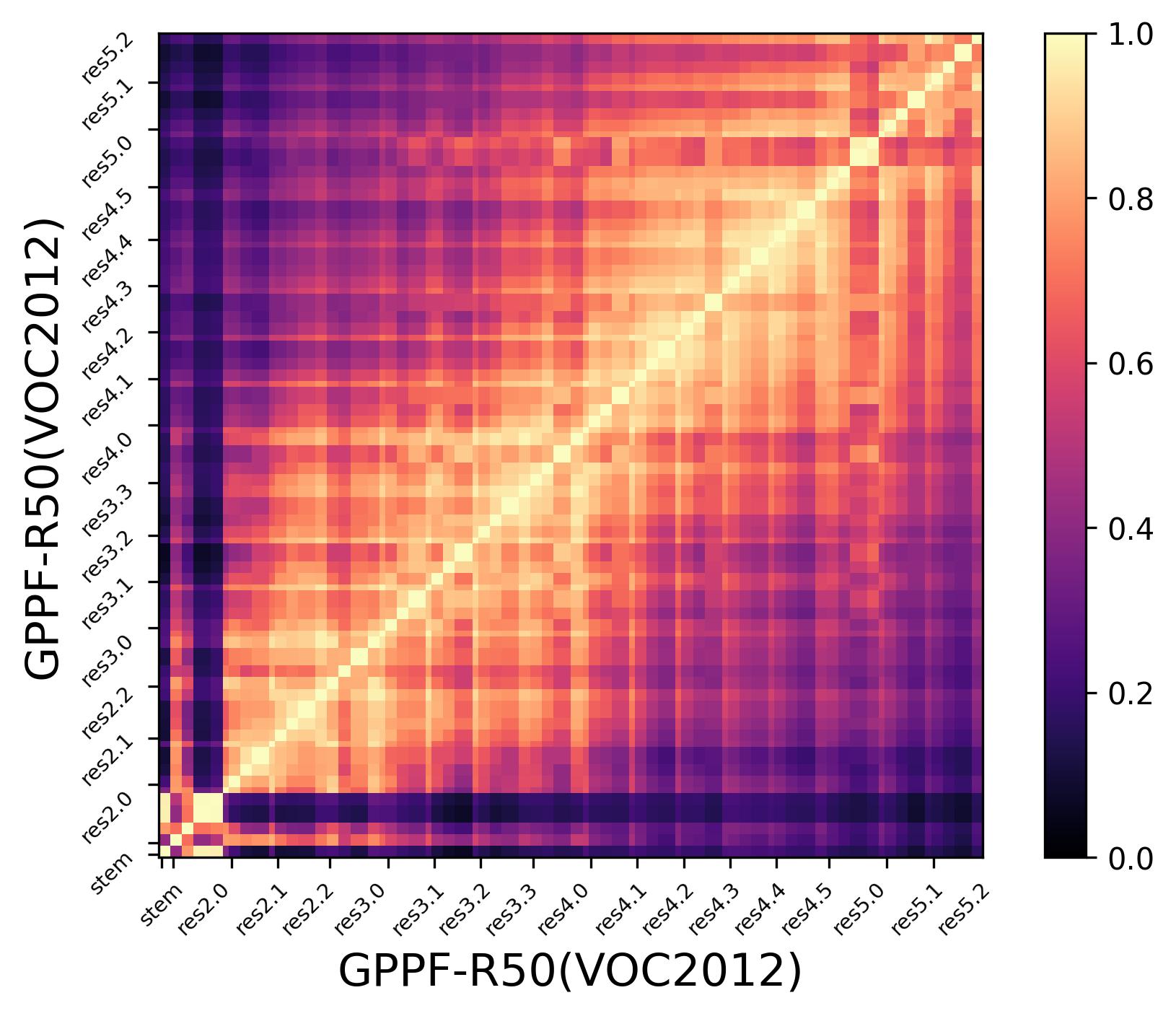}
		\caption{GPPF-R50}
	\end{subfigure}
	\hfill
	\begin{subfigure}{0.32\linewidth}
		\centering
		\includegraphics[width=\linewidth]{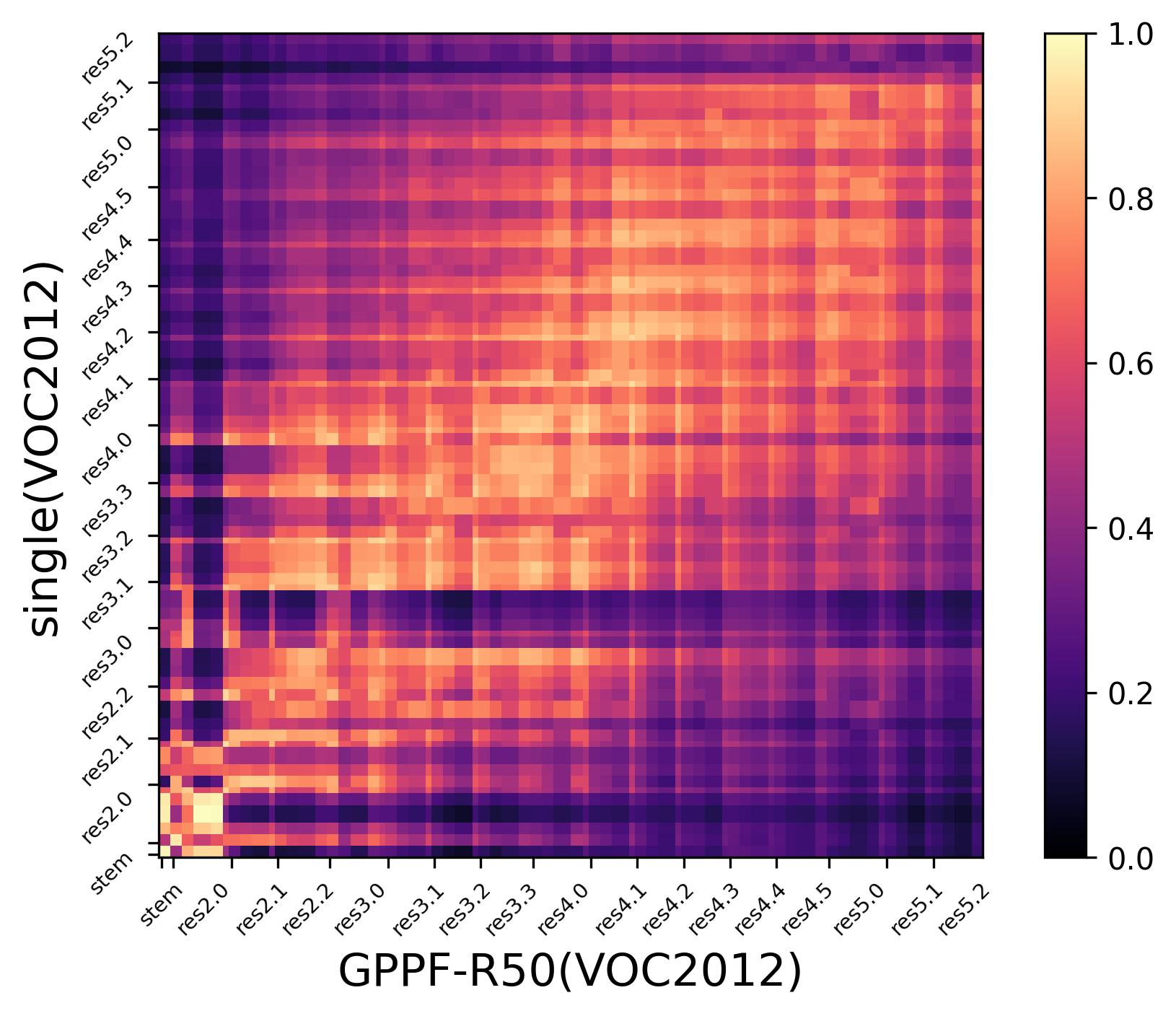}
		\caption{Cross Model Similarity}
	\end{subfigure}
	\caption{CKA analysis on Pascal VOC segmentation models.}
	\label{fig:voc2012}
\end{figure}

\begin{figure}[H]
	\centering
	\vspace{-15pt}
	\begin{subfigure}{0.32\linewidth}
		\centering
		\includegraphics[width=\linewidth]{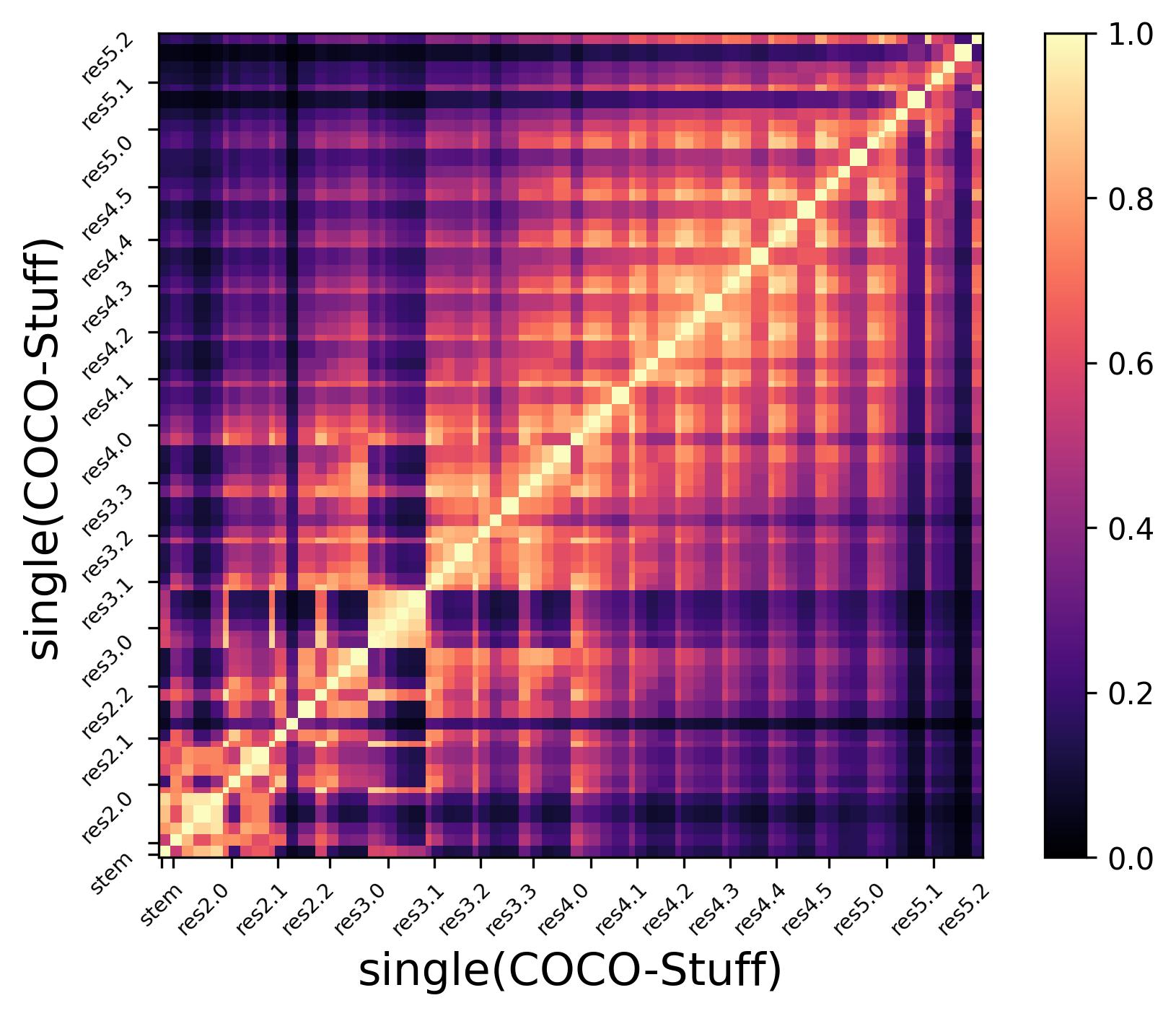}
		\caption{Single}
	\end{subfigure}
	\hfill
	\begin{subfigure}{0.32\linewidth}
		\centering
		\includegraphics[width=\linewidth]{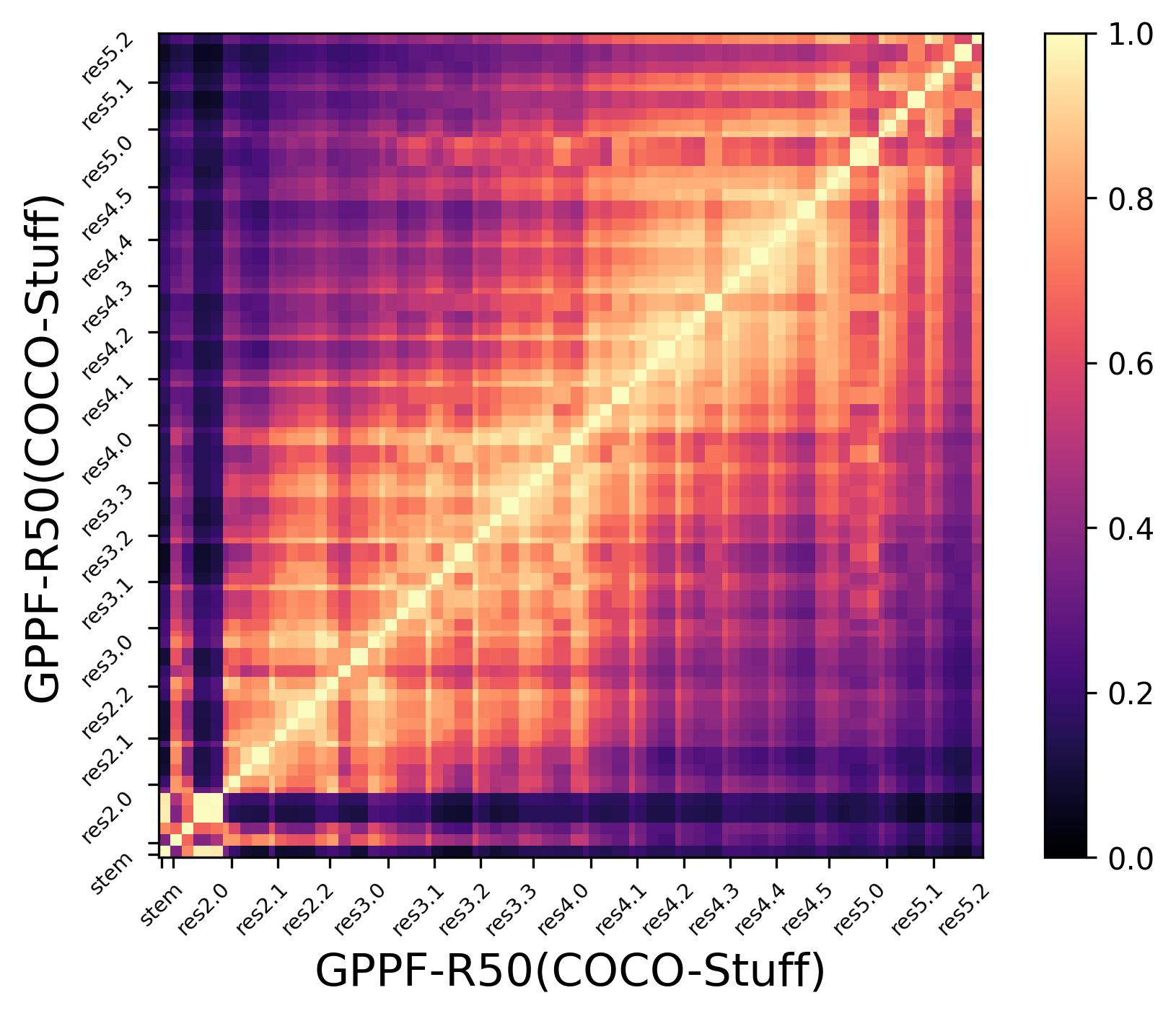}
		\caption{GPPF-R50}
	\end{subfigure}
	\hfill
	\begin{subfigure}{0.32\linewidth}
		\centering
		\includegraphics[width=\linewidth]{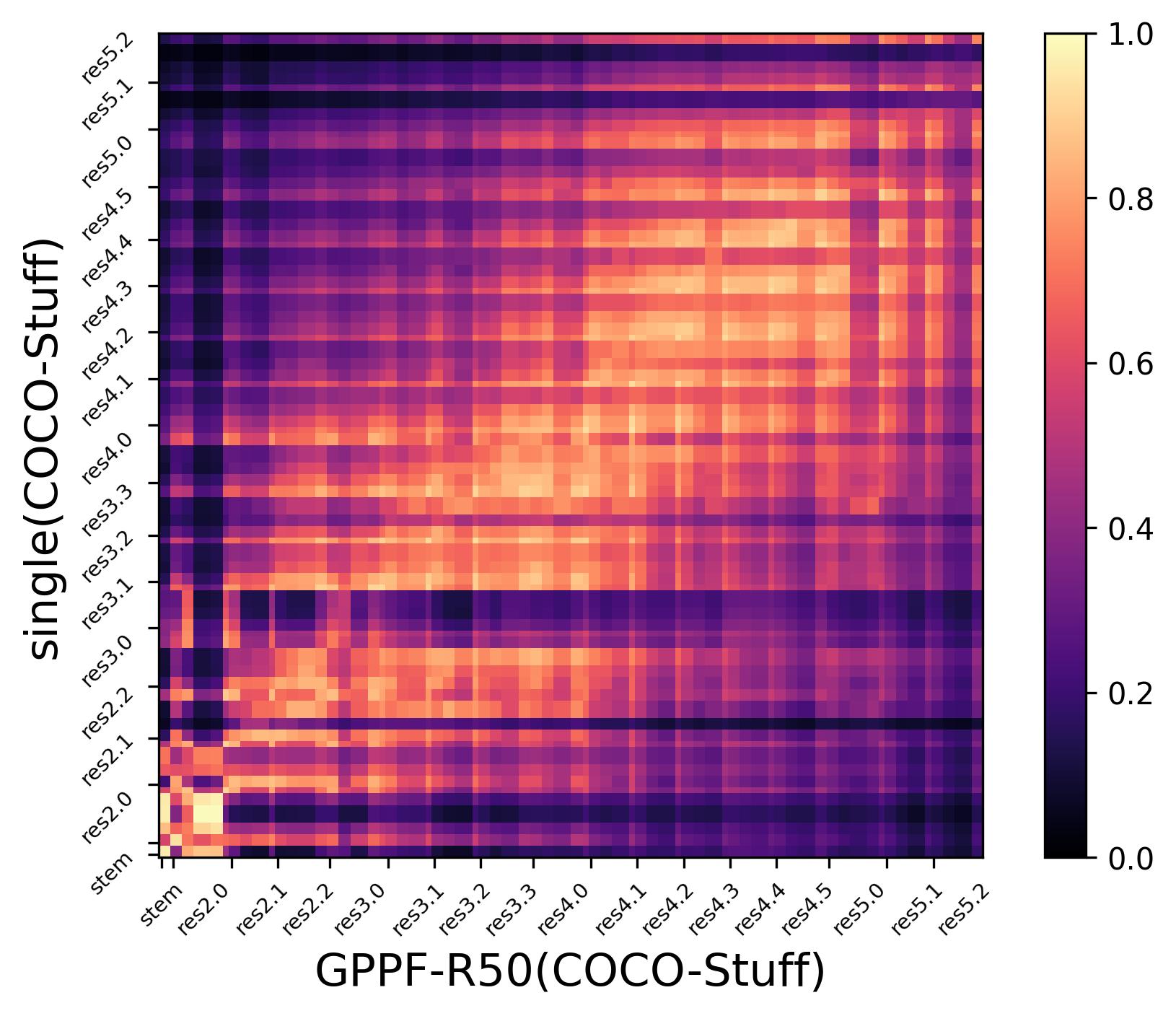}
		\caption{Cross Model Similarity}
	\end{subfigure}
	\caption{CKA analysis on COCO-Stuff segmentation models.}
	\label{fig:cocostuff}
\end{figure}

\begin{figure}[H]
	\centering
	\vspace{-15pt}
	\begin{subfigure}{0.32\linewidth}
		\centering
		\includegraphics[width=\linewidth]{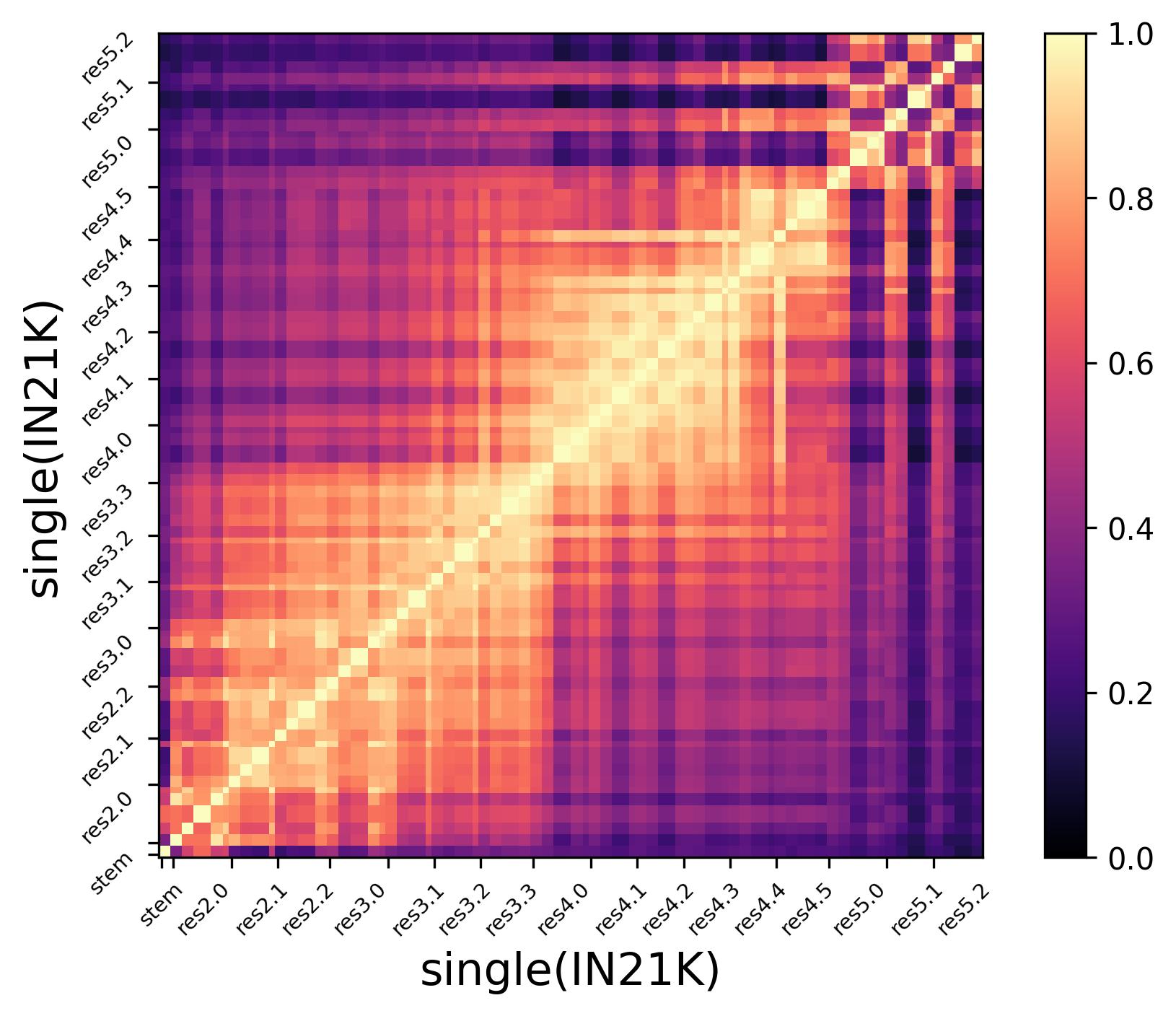}
		\caption{Single}
	\end{subfigure}
	\hfill
	\begin{subfigure}{0.32\linewidth}
		\centering
		\includegraphics[width=\linewidth]{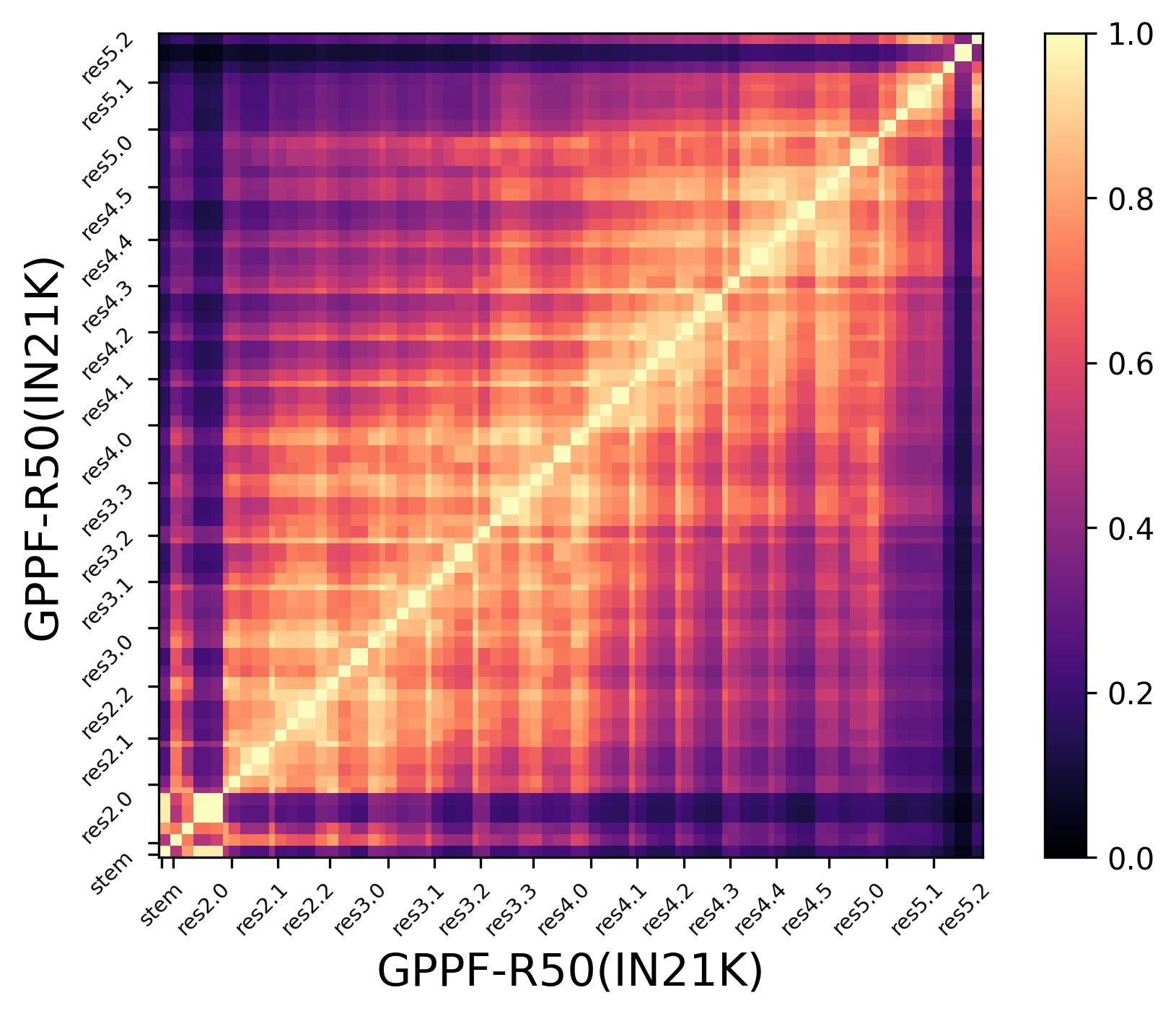}
		\caption{GPPF-R50}
	\end{subfigure}
	\hfill
	\begin{subfigure}{0.32\linewidth}
		\centering
		\includegraphics[width=\linewidth]{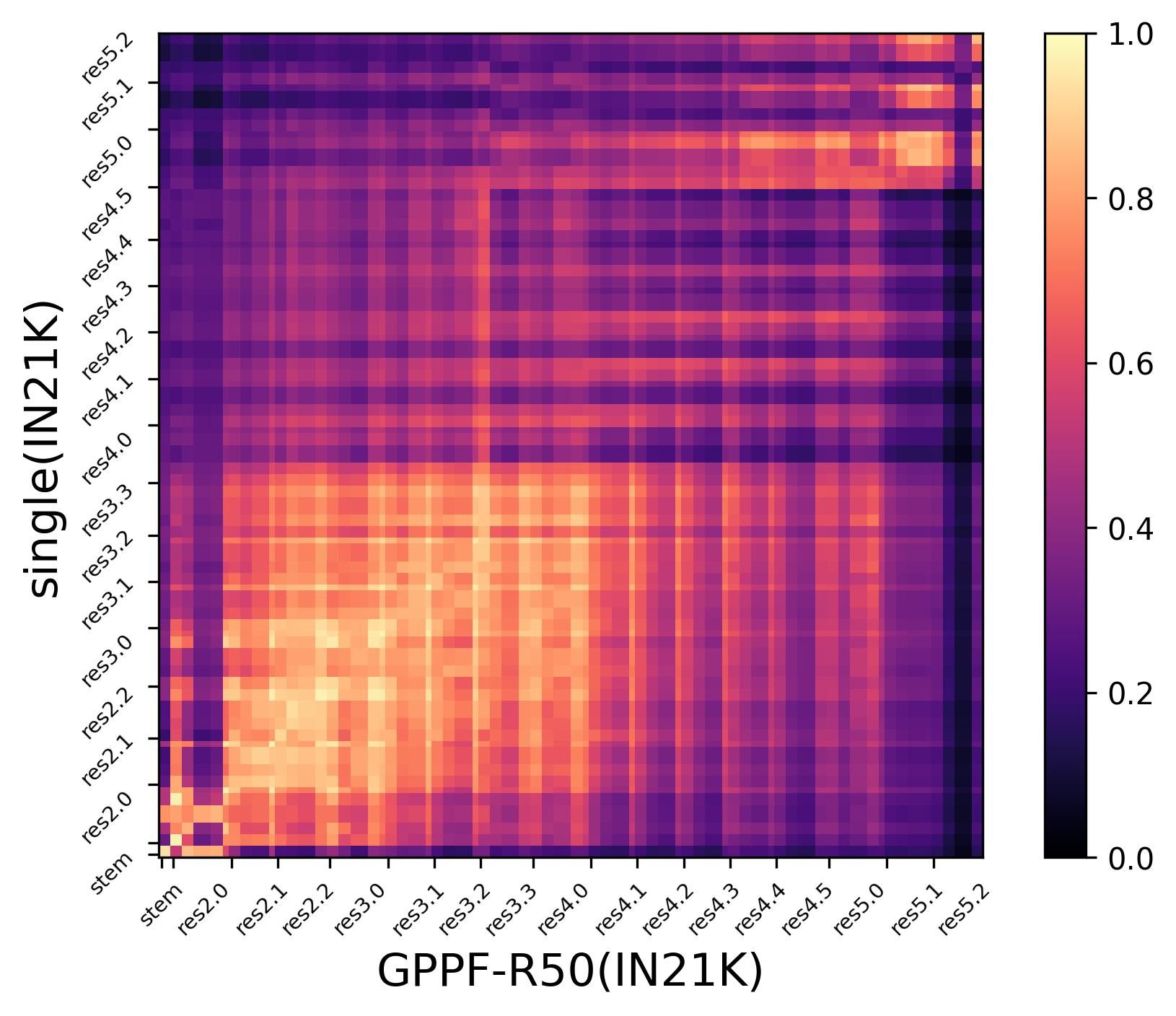}
		\caption{Cross Model Similarity}
	\end{subfigure}
	\caption{CKA analysis on ImageNet21K classification models.}
	\label{fig:in21k}
\end{figure}

\begin{figure}[H]
	\centering
	\vspace{-15pt}
	\begin{subfigure}{0.32\linewidth}
		\centering
		\includegraphics[width=\linewidth]{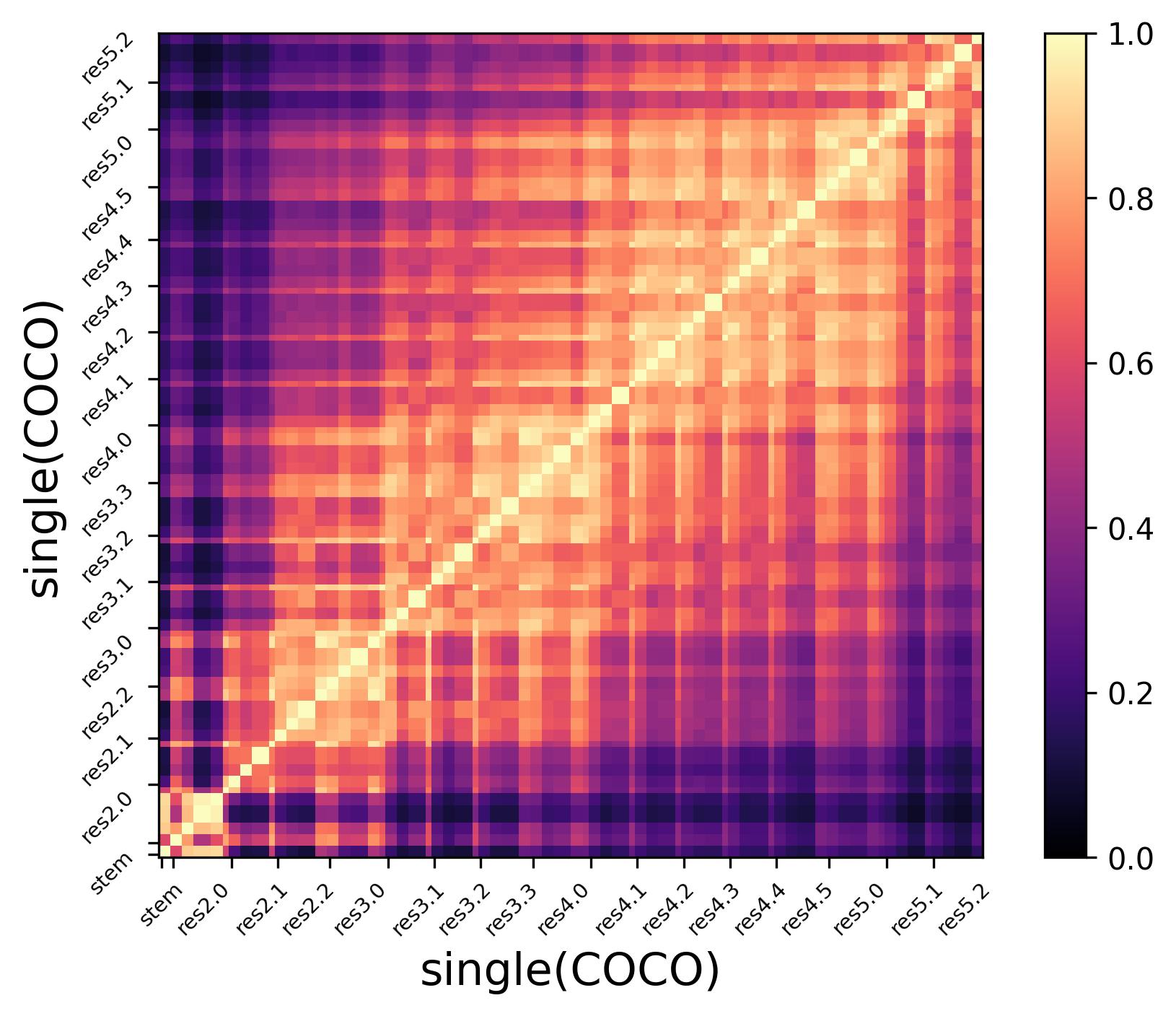}
		\caption{Single}
	\end{subfigure}
	\hfill
	\begin{subfigure}{0.32\linewidth}
		\centering
		\includegraphics[width=\linewidth]{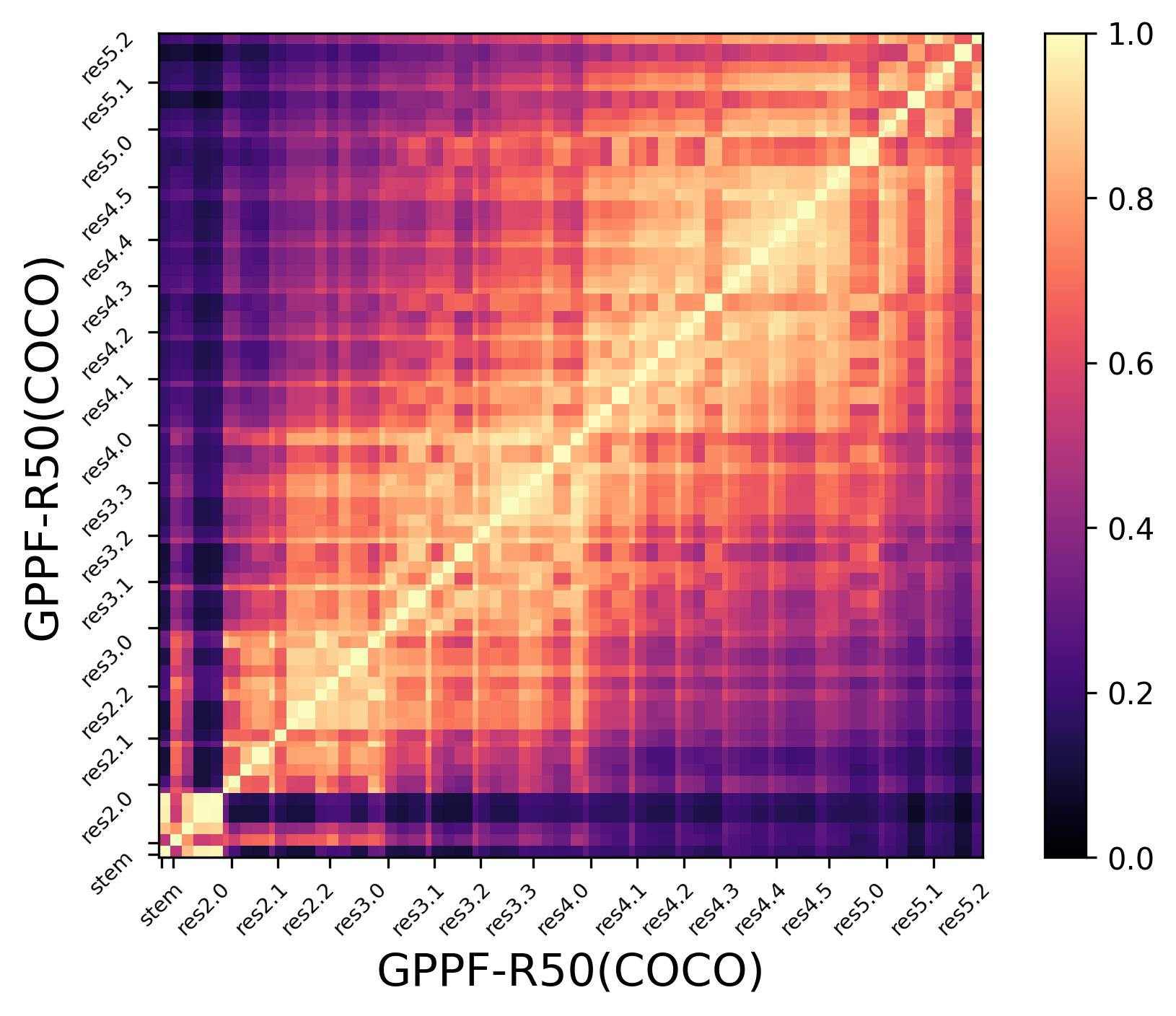}
		\caption{GPPF-R50}
	\end{subfigure}
	\hfill
	\begin{subfigure}{0.32\linewidth}
		\centering
		\includegraphics[width=\linewidth]{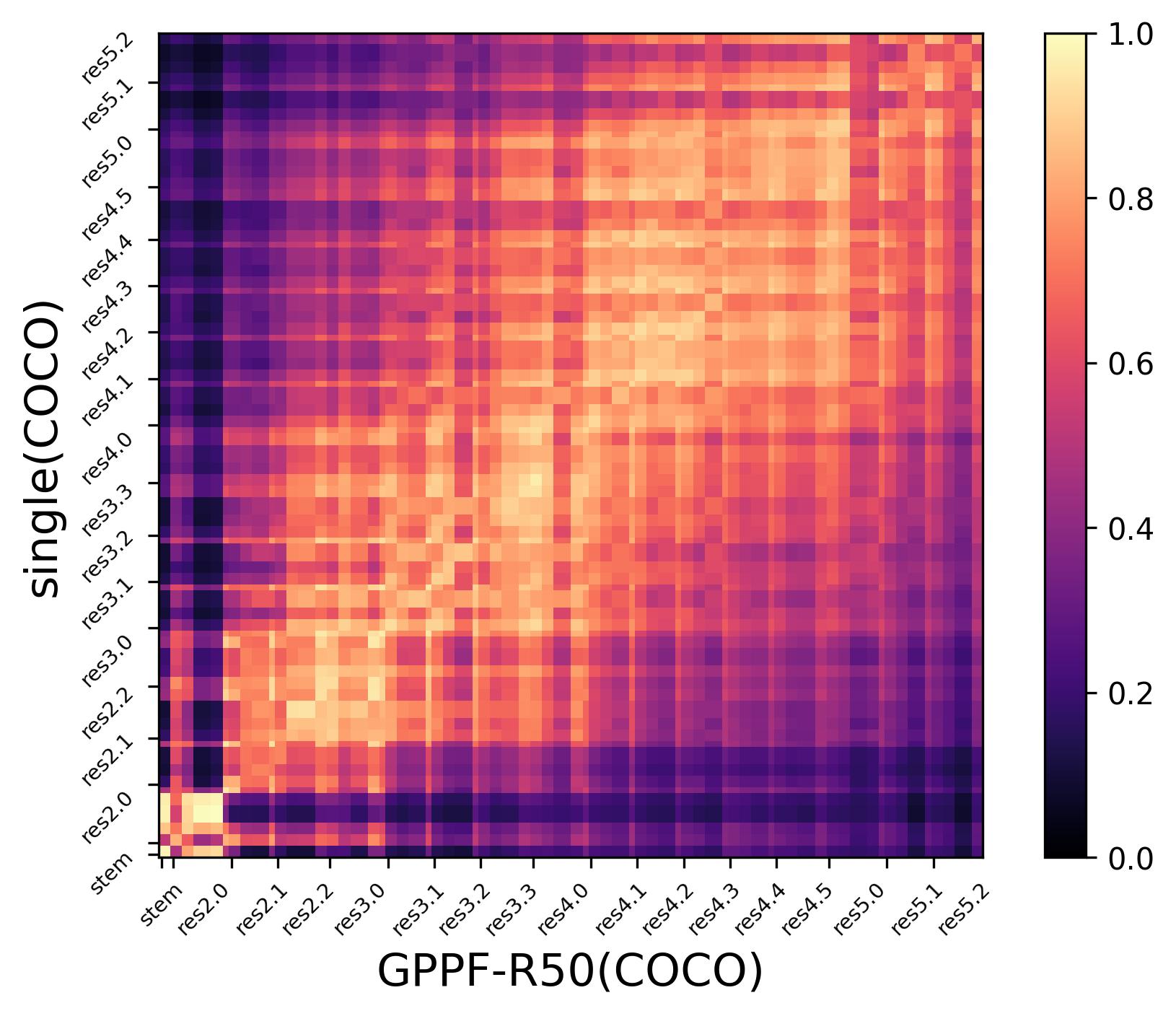}
		\caption{Cross Model Similarity}
	\end{subfigure}
	\caption{CKA analysis on COCO detection and instance segmentation models.}
	\label{fig:coco}
\end{figure}

\begin{figure}[H]
	\centering
	\vspace{-15pt}
	\includegraphics[width=0.6\linewidth]{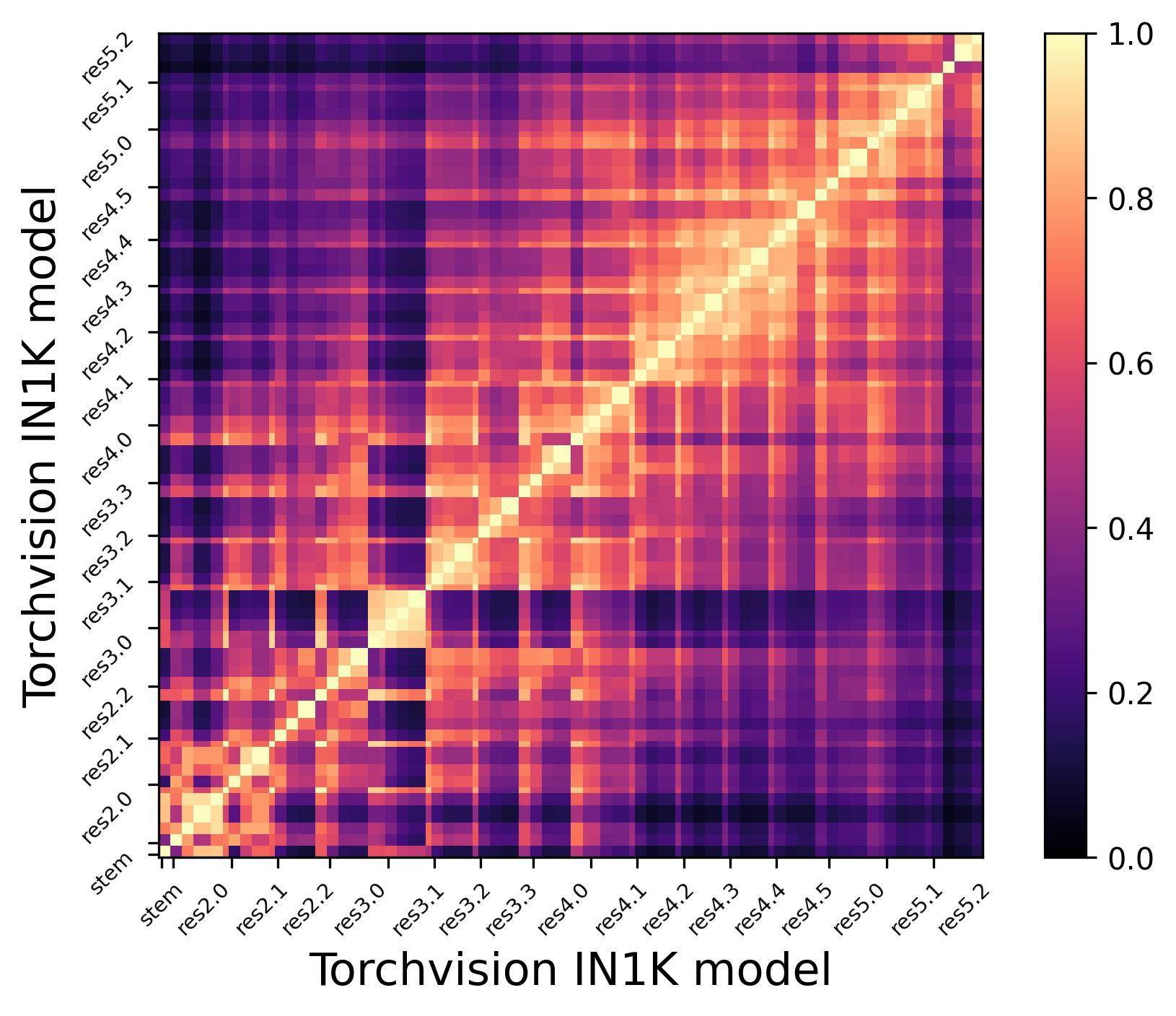}
	\caption{CKA analysis on Torchvision ImageNet1K pre-trained ResNet-50 model.}
	\label{fig:in1k}
\end{figure}

As demonstrated in the experiments, models trained by GPPF greatly outperforms the single task training. We wish to explore some of the beneficial changes by utilizing CKA analysis. For the three segmentation tasks, we observe the same phenomenon as depicted in Figure \ref{fig:ade20k}, Figure \ref{fig:voc2012}, and Figure \ref{fig:cocostuff}. In all the single task models, there is a clear "block structure" \citep{widedeep2021} between block "res3.1" and block "res3.2". This block structure is found related to prunable redundant networks  \citep{widedeep2021}, which is not useful for a deep model. On the other hand, the CKA analysis shows that GPPF-R50 does not have such defect in feature correlations. The correlation between layers is smoother, meaning the network has less redundant layers. By digging deeper into this, we discover the similar block structure in the pre-trained ResNet-50 model (See Figure \ref{fig:in1k}) given by TorchVision \footnote{\url{https://pytorch.org/vision/0.11/models.html\#torchvision.models.resnet50}}, which is used by the three segmentation tasks. We infer that the block structure in these three segmentation models are inherited from the ResNet-50 pre-train model. It has already been uncovered that finetuning can hardly escape from local minimum \citep{Neyshabur2020}. Similarly, we infer that finetuning can not esacape from the redundant block structure, meaning that it is hard to increase inforamtion in these layers. Therefore, having a better pre-training model becomes more important, since the initial CKA pattern largely decides the CKA pattern after finetuning. The figures show that our GPPF joint training gives a better initial point.  We also post the CKA analysis of ImageNet21K classification and COCO detection (and instance segmentation) in Figure \ref{fig:in21k} and Figure \ref{fig:coco}. Similar findings appear in ImageNet21K classification top layers. On the other hand, on COCO dataset, there is no block structure in the strong baseline of Copy-Paste \citep{copypaste2021ghiasi}, so the CKA analysis does not reflect big difference between single task training and joing training in the feature correlation.

\section{Analysis of Gumbel-Softmax Controller Learning}

\begin{figure}[H]
	\centering
	\hspace{40pt}
	\includegraphics[width=0.6\linewidth]{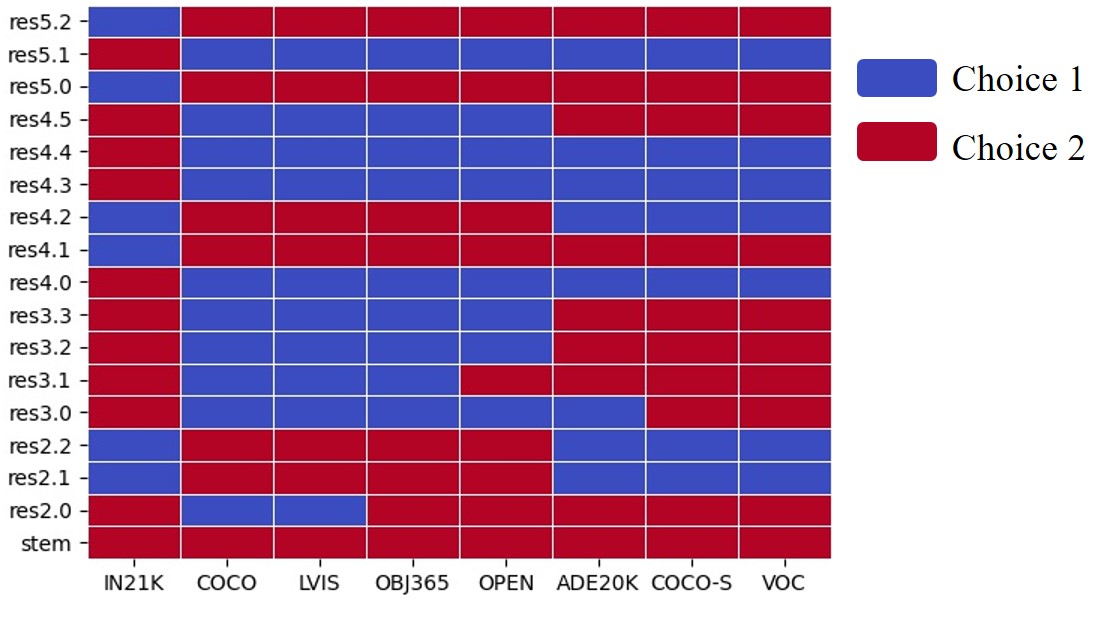}
	\caption{Lego Selection of GPPF-R50.}
	\label{fig:r50_choice}
\end{figure}

\begin{figure}[t]
	\centering
	\begin{subfigure}{0.3\linewidth}
		\centering
		\includegraphics[width=\linewidth]{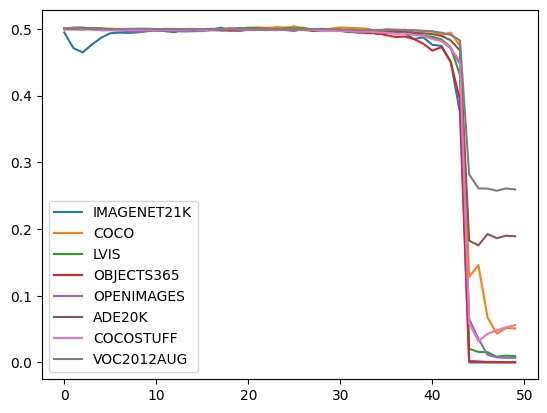}
		\caption{Stem}
	\end{subfigure}
	\hfill
	\begin{subfigure}{0.3\linewidth}
		\centering
		\includegraphics[width=\linewidth]{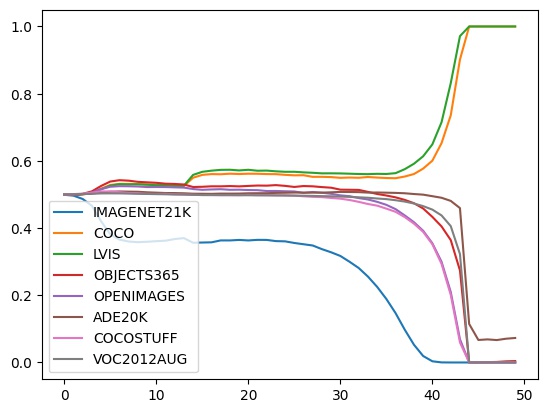}
		\caption{Res2.0}
	\end{subfigure}
	\hfill
	\begin{subfigure}{0.3\linewidth}
		\centering
		\includegraphics[width=\linewidth]{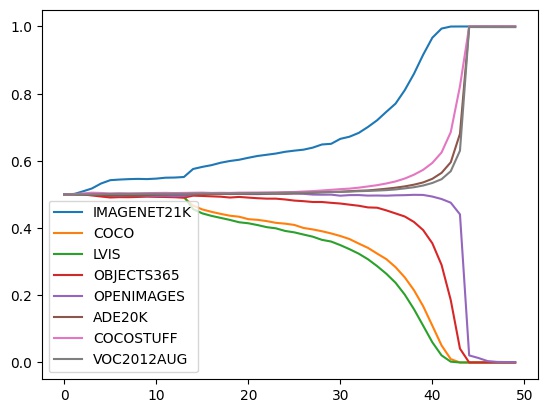}
		\caption{Res2.1}
	\end{subfigure}
	\hfill
	\begin{subfigure}{0.3\linewidth}
		\centering
		\includegraphics[width=\linewidth]{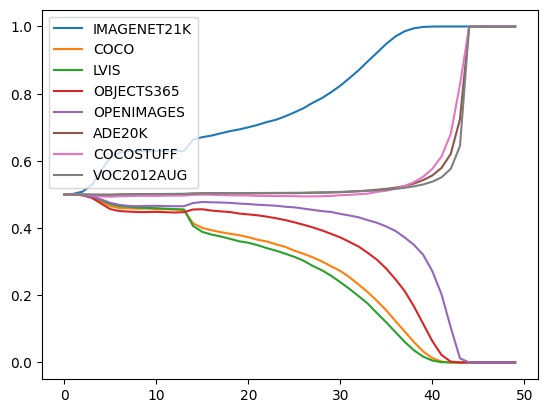}
		\caption{Res2.2}
	\end{subfigure}
	\hfill
	\begin{subfigure}{0.3\linewidth}
		\centering
		\includegraphics[width=\linewidth]{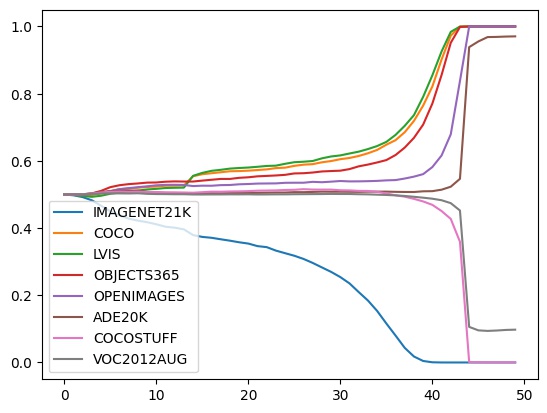}
		\caption{Res3.0}
	\end{subfigure}
	\hfill
	\begin{subfigure}{0.3\linewidth}
		\centering
		\includegraphics[width=\linewidth]{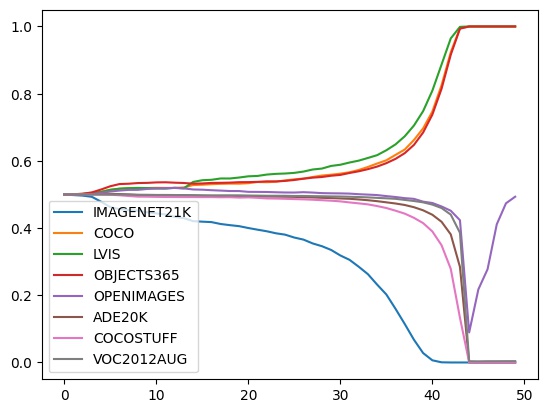}
		\caption{Res3.1}
	\end{subfigure}
	\hfill
	\begin{subfigure}{0.3\linewidth}
		\centering
		\includegraphics[width=\linewidth]{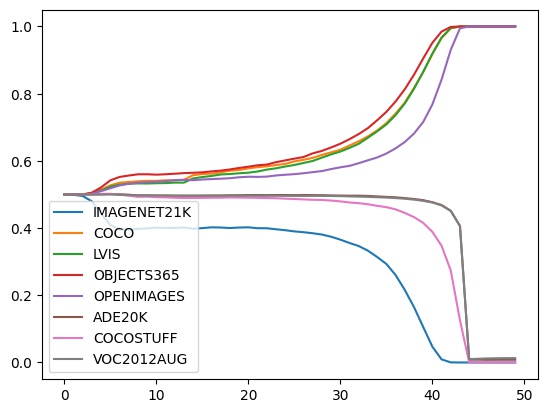}
		\caption{Res3.2}
	\end{subfigure}
	\hfill
	\begin{subfigure}{0.3\linewidth}
		\centering
		\includegraphics[width=\linewidth]{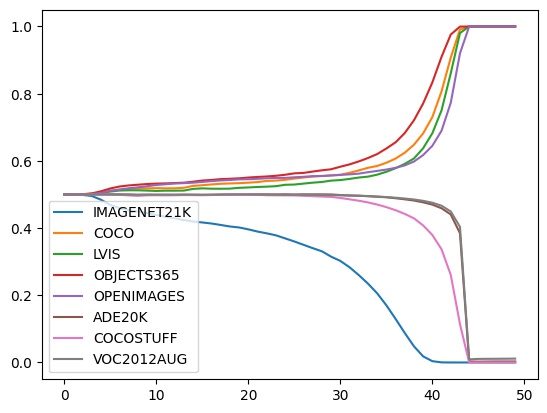}
		\caption{Res3.3}
	\end{subfigure}
	\hfill
	\begin{subfigure}{0.3\linewidth}
		\centering
		\includegraphics[width=\linewidth]{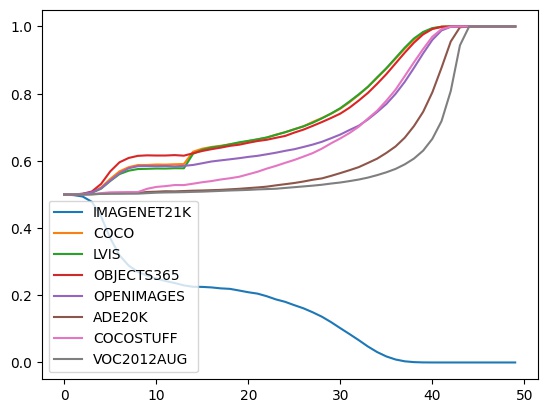}
		\caption{Res4.0}
	\end{subfigure}
	\hfill
	\begin{subfigure}{0.3\linewidth}
		\centering
		\includegraphics[width=\linewidth]{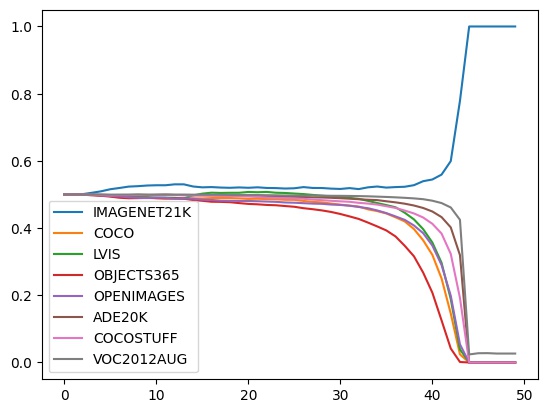}
		\caption{Res4.1}
	\end{subfigure}
	\hfill
	\begin{subfigure}{0.3\linewidth}
		\centering
		\includegraphics[width=\linewidth]{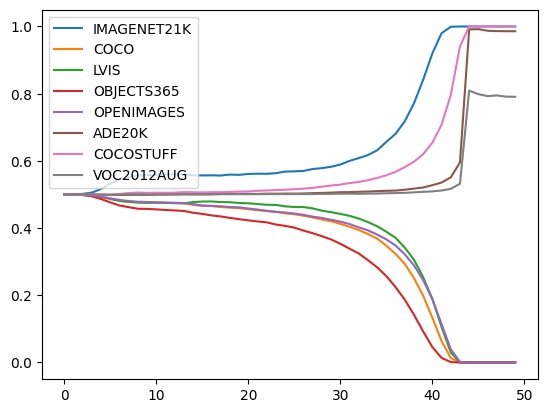}
		\caption{Res4.2}
	\end{subfigure}
	\hfill
	\begin{subfigure}{0.3\linewidth}
		\centering
		\includegraphics[width=\linewidth]{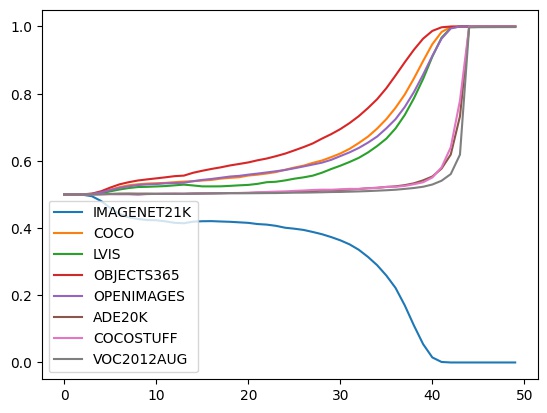}
		\caption{Res4.3}
	\end{subfigure}
	\hfill
	\begin{subfigure}{0.3\linewidth}
		\centering
		\includegraphics[width=\linewidth]{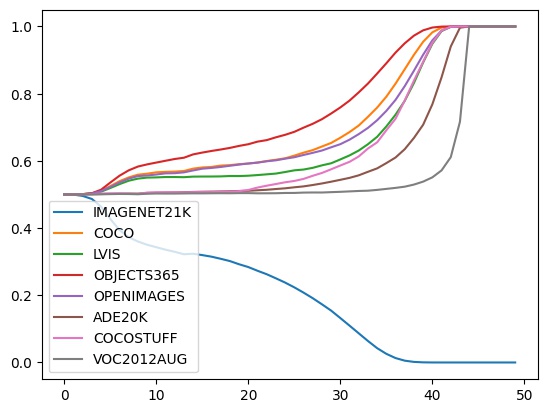}
		\caption{Res4.4}
	\end{subfigure}
	\hfill
	\begin{subfigure}{0.3\linewidth}
		\centering
		\includegraphics[width=\linewidth]{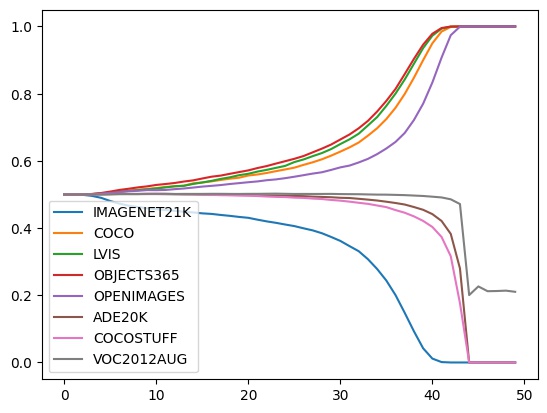}
		\caption{Res4.5}
	\end{subfigure}
	\hfill
	\begin{subfigure}{0.3\linewidth}
		\centering
		\includegraphics[width=\linewidth]{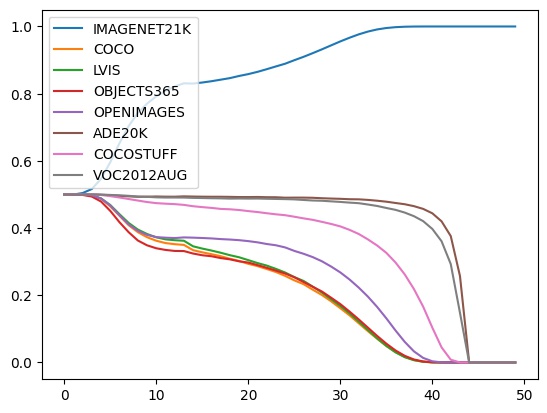}
		\caption{Res5.0}
	\end{subfigure}
	\hfill
	\begin{subfigure}{0.3\linewidth}
		\centering
		\includegraphics[width=\linewidth]{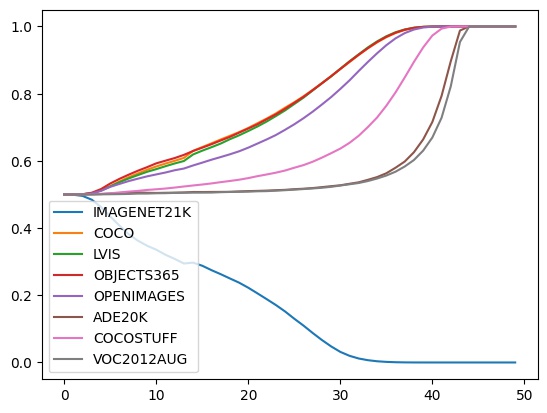}
		\caption{Res5.1}
	\end{subfigure}	
	\hspace{50pt}
	\begin{subfigure}{0.3\linewidth}
		\centering
		\includegraphics[width=\linewidth]{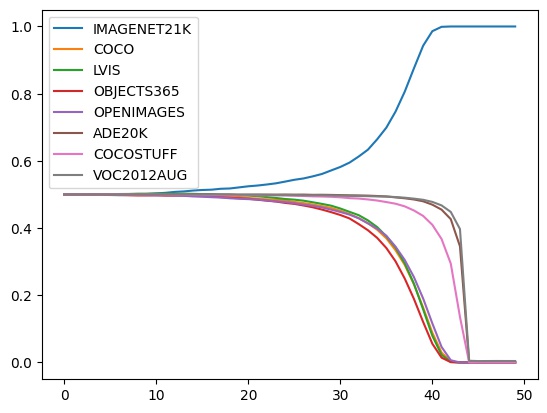}
		\caption{Res5.2}
	\end{subfigure}
	\caption{The selection learning process of each layer during training.}
	\label{fig:learning_process}
\end{figure}

For clarity, we include the lego selection of GPPF-R50 again in Figure \ref{fig:r50_choice}. To inspect how the selection probability changes during the training process, we plot the learning curves of each lego layer with an interval of 1w iterations in Figure \ref{fig:learning_process}. There is an acceleration of probability changes after 40w iterations caused by temperature decaying from around 0.6 to 0.01. For ImageNet21K, the probability converges quickly in most layers, showing that it has a clear difference/similarity with other tasks. This is also the case for COCO and LVIS, which trains both detection and instance segmentation. On the other hand, the three segmentation tasks maintain ambiguous selections until the quick drop of temerature in most layers. This might indicate that sharing with either classification or detection is beneficial for segmentation. The choice is hard to made until the temperature decay forces it to make one.
Finally, all tasks like to share the stem layer of ResNet. This also happens in the $N=3$ case and Swin-Transformer case. This implies that sharing the input processing layer is generally a good choice.

\clearpage

\section{Qualitative Results}

\subsection{Qualitative Results on KITTI depth estimation}
\begin{figure}[H]
	\centering
	\begin{subfigure}{0.9\linewidth}
		\centering
		\includegraphics[width=\linewidth]{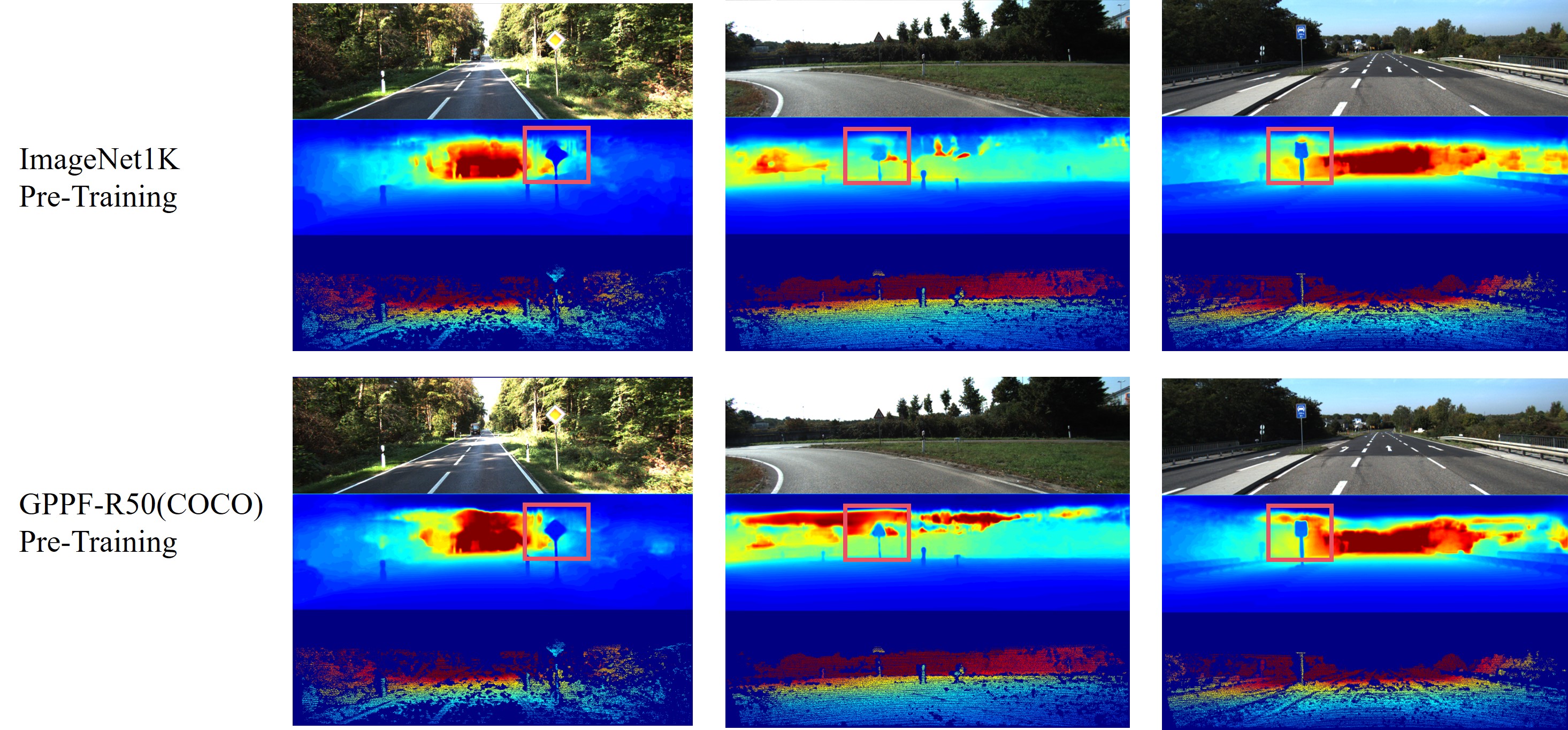}
		\caption{Improvements on sign depth estimation.}
		\label{fig:kitti_sign}
	\end{subfigure}
	
	\begin{subfigure}{0.9\linewidth}
		\centering
		\includegraphics[width=\linewidth]{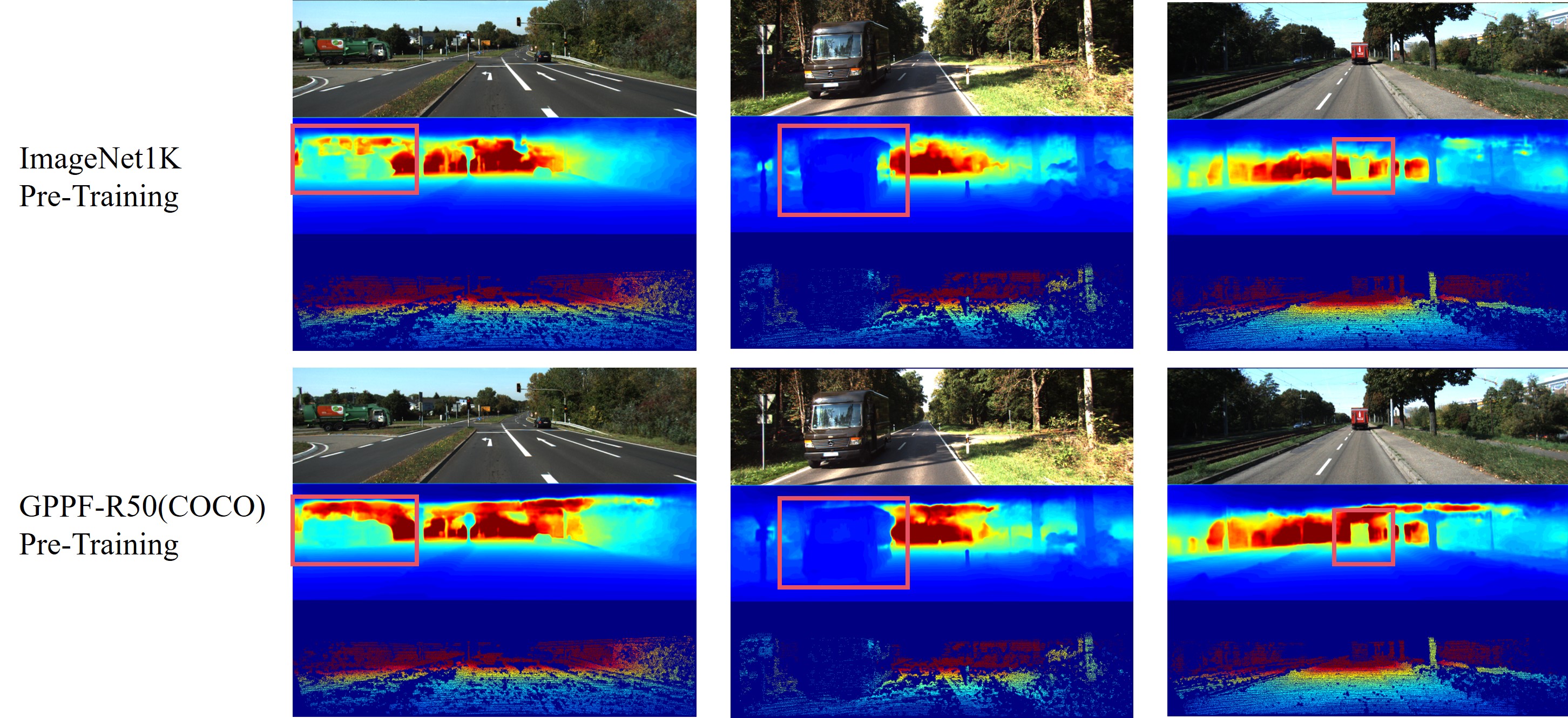}
		\caption{Improvements on car depth estimation.}
		\label{fig:kitti_car}
	\end{subfigure}
	\caption{Qualitative results on KITTI depth estimation. For each image, the sub-figures are original image, model prediciton, and ground truth from top to bottom. }
	\label{fig:kitti}
\end{figure}

We post some of the qualitative results of KITTI depth estimation in Figure \ref{fig:kitti}.  In Figure \ref{fig:kitti_sign}, we show some of the improvements on the depth estimation of traffic signs. It can be clearly observed that GPPF-R50 makes clearer and tighter depth prediciton on the sign area, while the ImageNet1K pre-traininig sometimes does not predict the boundary well. We also include some of the improvements on cars in Figure \ref{fig:kitti_car}. The ImageNet1K pre-trained model can not distinguish between the cars and background in some cases, while our GPPF-R50 pre-trained model predicts these regions of cars well. Both signs and cars have appeared in the object detetion and instance segmentation datasets like COCO. This proves that transferring from different sources is more helpful.

\section{Expermental Settings}
\label{section:exp}
\subsection{Pre-Training Hyperparameters}

We first listed the experimental setting for each single task. For IN21K classification, the hyperparmeter for Mixup, Cutmix, Random Erasing, and Label Smoothing is set as $0.8$, $1.0$, $0.25$, and $0.1$, respectively. For the 4 detection tasks, we choose Mask-RCNN \citep{maskrcnn} for COCO and LVIS and Faster R-CNN \citep{fasterrcnn} for Objects365 and Open Images. All the four detection tasks use large scale jitter with size 1024. Copy-Paste augmentation \citep{copypaste2021ghiasi} is  used for COCO and LVIS. For the segmentation tasks, we use UperNet \citep{xiao2018unified} with an auxiliary FCN head at the output of res4. Specifically, for the sementation single task baselines, we adopt ImageNet1K pre-training because training from scratch gets a low mIOU and cannot  reflect the relative improvements well. For the detection tasks, ImageNet1K pre-training is proved to underperform training from scratch in the selected training settings \citep{copypaste2021ghiasi}. So we report the performance of training from scratch for detection tasks.

For GPPF-R50 model, we use 7 servers with 8 NVIDIA V100 to train the model. We use SGD optimizer to train it for a total 50w iterations with initial learning rate as 0.16, weight decay as $5e-4$, and momentum as 0.9. Cosine learning rate decay is used with a warmup of 5w iters. The temperature for the Gumbel-Softmax sampling is set as 5.0 at the start of the training and linearly decays to 0.01 at 45w iters. The temperature is kept fixed as 0.01 for the last 5w training iters. The task sampling weight (used to sample task on each GPU) of IN21K, COCO, LVIS, Objects365, Open Images, ADE20K, COCO-STUFF, and Pascal VOC is manually set to 4.0, 2.5, 2.5, 4.0, 4.0, 1.0, 2.0, and 1.0, repectively. The batch size on each GPU for each task is set as 96, 8, 8, 12, 12, 8, 8, and 8, respectively. Model Checkpoint is used, which greatly lowers the training. The total training costs about 2 weeks. For GPPF-SwinT, the training settings are similar, except we use AdamW optimizer with initial learning rate as 0.0001 and weight decay as 0.05. We also use Cascade R-CNN \citep{cascadercnn} for 4 detection tasks in the GPPF-SwinT model.

\subsection{Downstream Tasks}

\paragraph{Depth Estimation} For depth estimation on KITTI dataset, we use BTSHead \citep{lee2019big}. We select batch size as 32, loss weight as 0.1, and initial learning rate as 0.01. The parameters are optimized by SGD with weight decay as $1e-8$, momentum as 0.9, and total steps as 30000. Consine learning rate decay with a warmup of 1000 iters are used. We also use a stochastic depth  \citep{densenet2017huang} with probability of 0.4 in the finetuning.

\paragraph{Detection} The dataset-specific hyperparameters we used in UODB finetuning is listed in Table \ref{table:uodb_hyper}. The batch size for training is set as 16 for all datasets. The training steps are mostly decided by its cardinality except for datasets like KITTI, where we observe an obvious overfitting if we use long training steps. For anchor sizes, we keep the same as COCO except for WiderFace, LISA, and DeepLesion. These three datasets contain many small objects and we half the original anchor sizes for each FPN layer. The initial learning rate for finetuning is set as 0.01 and decays two times with a factor of 0.1. The optimizer is chosen as SGD with momentum 0.9. We also use a weight decay of $4e-5$ as in the pre-training stage.

\begin{table}[h]
	\caption{The hyperparamters of UODB finetuning.}
	\label{table:uodb_hyper}
	\begin{tabular}{l|cccc}
		\toprule
		Dataset & Max Steps & Decay Steps &  Test Size & Anchor Size \\
		\midrule
		KITTI 		& 4000 & 3000,3500 & 800 & [[32], [64], [128], [256], [512]] \\
		WiderFace 	& 26000 & 18000,24000 & 800 & [[16], [32], [64], [128], [256]] \\
		PascalVOC 	& 18000 & 12000,16000 & 800 & [[32], [64], [128], [256], [512]] \\
		LISA 		& 18000 & 12000,16000 & 800 & [[16], [32], [64], [128], [256]] \\
		DOTA 		& 26000 & 18000,24000 & 800 & [[32], [64], [128], [256], [512]] \\
		Watercolor  & 1800  & 1500,1700   & 800 & [[32], [64], [128], [256], [512]] \\
		Clipart     & 1800  & 1500,1700   & 800 & [[32], [64], [128], [256], [512]] \\
		Comic       & 1800  & 1500,1700   & 800 & [[32], [64], [128], [256], [512]] \\
		Kitchen 	& 13500 & 9000,12000  & 800 & [[32], [64], [128], [256], [512]] \\
		DeepLesion  & 50000 & 30000,40000 & 512 & [[16], [32], [64], [128], [256]] \\
		\bottomrule
	\end{tabular}
	
\end{table}

\paragraph{Classification} For ImageNet1K single task training and finetuning, we use the same hyperparameters except for learning rate. Similar to the ImageNet21K pre-training, the hyperparmeter for Mixup, Cutmix, Random Erasing, and Label Smoothing is set as $0.8$, $1.0$, $0.25$, and $0.1$, respectively. We totally train the model for  300 epochs. The optimizer is chosen as SGD with momentum and weight decay set as 0.9 and $1e-4$. The initial learning rate for training from scratch is 0.1 and the finetuning learning rate is 0.01. Cosine learning rate decay is used with a warmup of 20 epochs. The batch size used for training is 2048. For DyFinetune, we use half training steps to decay the temperature from 5.0 to 0.01 and keep it fixed for the rest steps.

For the other downstream classification datasets, we follow the protocols in previous works \citep{byol2020,simclr2020}. We only apply random crops and random flip during training. At test time, the short side of image is resized to 256 and followed by a center crop of $224\times 224$. We use SGD optimizer to train for a total 20000 iters for all datasets with a batch size of 256 and momentum of 0.9. For learning rate and weight decay, a grid search is applied to search the best combination. The search space is 7 logarithmically spaced between n 0.0001 and 0.1 for learning  rate and 7 logarithmically-spaced values of weight decay between $1e-6$ and $1e-3$, as well as 0 for weight decay. Weight decay will be divided by learning rate.

\section{Asset License and Term of Use}

The assets used in the paper is listed here (we do not include those we can not find the license):

\begin{itemize}
	\item \textbf{ImageNet}:  Custom (research, non-commercial)
	\item \textbf{COCO}: Creative Commons Attribution 4.0 License
	\item \textbf{LVIS}: Creative Commons Attribution 4.0 License
	\item \textbf{Objects365\_v2}: Creative Commons Attribution 4.0 License
	\item \textbf{Open Images}: CC BY 4.0
	\item \textbf{ADE20K}: Creative Commons BSD-3 License Agreement
	\item \textbf{Pascal VOC 2012}: "flickr" terms of use
	\item \textbf{COCO-Stuff}: Creative Commons Attribution 4.0 License
	\item \textbf{KITTI}: CC BY-NC-SA 3.0
	\item \textbf{DOTA}: "Google Earth" terms of use \& Custom (non-commercial)
	% \item \textbf{DeepLesion}: Unknown
	\item \textbf{Clipart,Comic,Watercolor}: Unknown
	\item \textbf{LISA}: CC BY-NC-SA 4.0
	% \item \textbf{Kitchen}: Unknown
	\item \textbf{Aircraft}: Custom (non-commercial)
	\item \textbf{Cars}: Custom (non-commercial)
	\item \textbf{Pets}: CC BY-SA 4.0
	\item \textbf{PyTorch}: \url{https://github.com/pytorch/pytorch/blob/master/LICENSE}
	\item \textbf{MMSegmentation}: Apache License Version 2.0
	\item \textbf{Detectron2}: Apache License Version 2.0
\end{itemize}

\end{document}